\documentclass[12pt]{article}

\usepackage[utf8]{inputenc}
\usepackage[T1]{fontenc}
\usepackage[authoryear,round]{natbib}
\usepackage{amsmath,amssymb,graphicx}
\usepackage{amsfonts,amsthm}
\graphicspath{{./}{figures/}}
\usepackage[linesnumbered,ruled,vlined]{algorithm2e}
\usepackage{xcolor}
\usepackage{booktabs}
\usepackage{multirow}
\usepackage{siunitx}
\newtheorem{remark}{Remark}

\usepackage{caption}

\usepackage{hyperref}
\hypersetup{
  pdftitle={ALMAB-DC: An Asynchronous Gaussian Process--Bandit Framework for
    Scalable Sequential Experimental Design},
  pdfauthor={Foo Hui-Mean; Yuan-chin Ivan Chang},
  pdfkeywords={response surface methodology; Bayesian optimal design;
    dose--response optimization; adaptive spatial sampling;
    multi-armed bandits; asynchronous distributed computing},
  colorlinks=true,
  linkcolor={blue},
  filecolor={blue},
  citecolor={blue},
  urlcolor={blue}}

\newcommand{\anon}{1}

\begin{document}

\def\spacingset#1{\renewcommand{\baselinestretch}%
{#1}\small\normalsize} \spacingset{1}


\if1\anon
{
  \title{\bf ALMAB-DC: An Asynchronous Gaussian Process--Bandit Framework for
    Scalable Sequential Experimental Design}
  \author{Foo Hui-Mean\thanks{huimean87@gmail.com}\hspace{.2cm}\\
    Institute of Statistical Science, Academia Sinica, Taipei, Taiwan\\
    and \\
    Yuan-chin Ivan Chang\thanks{Corresponding author: ycchang@stat.sinica.edu.tw} \\
    Institute of Statistical Science, Academia Sinica, Taipei, Taiwan}
  \date{}
  \maketitle
} \fi

\if0\anon
{
  \bigskip
  \bigskip
  \bigskip
  \begin{center}
    {\LARGE\bf ALMAB-DC: An Asynchronous Gaussian Process--Bandit Framework for
    Scalable Sequential Experimental Design}
  \end{center}
  \medskip
} \fi

\begin{abstract}
Batch-synchronous experimental design wastes parallel computing resources by forcing workers to idle while waiting for the slowest evaluation to return. We propose ALMAB-DC (Active Learning, Multi-Armed Bandit, Distributed Computing), an asynchronous Gaussian process-bandit framework that eliminates this bottleneck. The framework integrates a Gaussian process surrogate, a multi-armed bandit allocation layer, and an event-driven distributed execution layer within a single sequential decision loop. A three-term regret decomposition characterizes each component's contribution, while scaling-law arguments provide guidance for worker-count selection. We validate the framework across three statistical design settings—response surface optimization, Bayesian dose–response design, and adaptive spatial sampling—demonstrating parallel scalability on computational benchmarks and transferability to pharmacogenomic and geostatistical datasets.
\end{abstract}

\vfill

\newpage
\spacingset{1}

\section{Introduction}
\label{sec:intro}

Sequential experimental design---the adaptive, budget-constrained allocation of
evaluation effort to maximize information gain or minimize a design
criterion---is fundamental to process engineering, pharmacology, spatial
science, and machine learning. Classical approaches, including Bayesian optimal
design \citep{chaloner1995bayesian}, D-optimal design, and space-filling
strategies, typically commit evaluation locations in advance or in small
synchronous batches. As parallel computational resources become available, a
structural bottleneck emerges: batch-synchronous frameworks force every worker
to idle until the slowest evaluation in the current batch returns before any new
candidate can be dispatched. This synchronization overhead scales with batch
size and is not merely an implementation inconvenience; it is a direct loss of
wall-clock efficiency that compounds as the number of parallel workers grows.

We therefore propose \textbf{ALMAB-DC} as an architectural solution to this 
bottleneck. ALMAB-DC is an event-driven asynchronous framework in which each 
completed evaluation immediately triggers a surrogate update and the dispatch 
of a new candidate, without any global synchronization barrier. The framework 
integrates three tightly coupled components. First, a \emph{Gaussian process 
surrogate} provides uncertainty-aware predictions of the unknown response or 
utility surface, while posterior-utility acquisition functions (UCB, EI, 
Max-Variance, and Thompson Sampling) translate this uncertainty into ranked 
candidate queries. Second, a \emph{multi-armed bandit layer} governs the 
allocation of evaluation effort across parallel workers and candidate regions, 
balancing exploration and exploitation with logarithmic regret guarantees. 
Third, a \emph{distributed asynchronous execution layer} built on Ray 
\citep{moritz2018ray} removes the synchronization barrier through a 
\texttt{WaitForAny} dispatch rule: as soon as any worker returns, its result is 
incorporated and the worker is immediately redeployed using Kriging Believer 
heuristics to maintain batch diversity.

In this paper, we demonstrate that the ALMAB-DC architecture extends naturally,
without modification, to two complementary evaluation contexts. The
\emph{primary} context is classical statistical experimental design:
sequential response surface optimization, Bayesian dose--response design for
clinical trials, and adaptive spatial sampling for Gaussian process field
estimation. Because these problems have historically been approached through
synchronous, fixed-batch designs, the asynchronous layer introduced here acts
as a practical design enabler, converting parallel computational resources
into near-linear reductions in wall-clock rounds while preserving---and in some
cases improving---the statistical objectives of interest. The \emph{secondary}
context is computational black-box optimization, represented by deep-learning
hyperparameter tuning, CFD drag minimization, and reinforcement-learning
policy search. These benchmarks play a supporting role, providing evidence
that the same architecture scales effectively and performs well relative to
strong baselines in engineering and machine-learning settings.

Existing approaches address complementary aspects of this challenge. Bayesian
optimization methods such as Optuna (TPE) carry a probabilistic surrogate but
operate synchronously and lack an explicit bandit allocation policy. BOHB and
parallel BO add batch parallelism but impose synchronization barriers and do not
incorporate bandit-driven allocation. Distributed MAB methods eliminate
synchronization but lack the surrogate that gives GP-based acquisition its
sample-efficiency advantage. No single framework simultaneously combines a GP
surrogate, adaptive bandit allocation, asynchronous execution, and statistical
design criteria. ALMAB-DC is designed to fill this gap.

\paragraph{Relation to asynchronous and batched Bayesian optimisation.}
Parallelising Bayesian optimisation has been studied through several
complementary lenses.
\citet{desautels2014parallelizing} introduced GP-BUCB and GP-AUCB, which
parallelise UCB-based acquisition using hallucinated observations and provide
the first regret bounds for parallel GP bandits; ALMAB-DC differs in adding an
explicit region-level bandit allocation layer that governs \emph{which part of}
$\mathcal{X}$ receives evaluation effort, rather than relying on hallucination
alone to discourage clustering.
\citet{alvi2019asynchronous} proposed asynchronous local penalisation (ALP),
which adds a repulsion term to the acquisition function to spread pending
queries; ALMAB-DC uses Kriging Believer variance deflation
\citep{ginsbourger2010kriging} for the same within-batch purpose and
additionally employs a region-UCB policy to route workers to distinct spatial
regions.
\citet{death2021asynchronous} proposed AEGiS, an asynchronous $\varepsilon$-greedy
BO method; ALMAB-DC replaces $\varepsilon$-greedy allocation with UCB-1 region
selection, recovering logarithmic cumulative allocation regret rather than
linear worst-case cost.
\citet{kandasamy2018parallelised} parallelised Thompson sampling for GP bandits;
ALMAB-DC supports Thompson sampling as an acquisition function and additionally
operates in the asynchronous regime, where workers need not synchronise.
None of these methods is designed for the statistical experimental design
objectives---integrated posterior variance minimisation, Bayesian utility
maximisation, and D-optimality---that are the primary focus here.
The Kriging Believer (KB) heuristic \citep{ginsbourger2010kriging} is
well established and implemented in libraries such as BoTorch.
The novelty of ALMAB-DC is not KB in isolation but its integration within an
asynchronous dispatch loop governed by a region-level UCB-1 allocation policy
applied to classical experimental design objectives; the ablation study in
Section~\ref{sec:ablation} isolates each component's contribution.

\paragraph{Contributions.}
ALMAB-DC makes three methodological contributions relevant to statistical
practice.
\textit{First}, it provides a unified sequential design loop that
simultaneously satisfies classical statistical objectives---simple regret
minimization, Bayesian utility maximization, and integrated posterior variance
reduction---across response surface, dose--response, and spatial sampling
settings without architectural modification.
\textit{Second}, it introduces a three-term regret decomposition
($R_T^{\mathrm{AL}}+R_T^{\mathrm{MAB}}+R_T^{\mathrm{async}}$)
as a conceptual organising framework that separates the statistical costs of
surrogate learning, region-bandit allocation, and asynchronous execution; each
term is bounded individually using established results
\citep{srinivas2010gaussian,auer2002finite,joulani2013online}, and a
practically actionable worker threshold
$K^\dagger=\mathcal{O}((T/\bar{C})^{1/2})$ characterizes the
efficiency--throughput trade-off.
\textit{Third}, it demonstrates that event-driven asynchronous
dispatch---via the \texttt{WaitForAny} rule with Kriging Believer batch
diversity---preserves statistical design objectives while delivering
near-linear wall-clock speedups, establishing that computational and
statistical efficiency are complementary, not competing, in sequential
experimental design.

The remainder of the paper is organized as follows.
Section~\ref{sec:framework} presents the full ALMAB-DC framework: problem
setup and statistical objective, GP surrogate with acquisition functions,
bandit allocation with Kriging Believer batch diversity, the regret
decomposition with scalability analysis, and practical implementation notes.
Section~\ref{sec:results} reports the empirical results: the evaluation
protocol (Section~\ref{sec:eval-protocol}), the three primary statistical
design applications---Cases~P1--P3 covering sequential response surface
design, Bayesian dose--response design, and adaptive spatial
sampling---followed by the secondary computational validation benchmarks
(Cases~C1--C3), ablation and scaling analyses, and external transferability
checks on pharmacogenomic and geostatistical public datasets.
Section~\ref{sec:discussion} discusses the main findings, including
validation of the regret decomposition and practical implications.
Section~\ref{sec:conclusion} concludes.

\section{Structure of ALMAB-DC}
\label{sec:framework}

\subsection{Statistical Objective}
\label{sec:setup}

Let $\mathcal{X}$ denote the candidate design space, which may be a grid, a
continuous region, or a finite arm set depending on the application. Let
$f:\mathcal{X}\to\mathbb{R}$ denote an unknown response or utility surface, and
suppose that observations take the form
$Y(x)=f(x)+\varepsilon(x)$, where
$\varepsilon(x)\sim\mathcal{N}(0,\sigma_n^2)$ is i.i.d.\ observation noise.
Because each evaluation is assumed to be expensive, the total evaluation
budget, say $T$, must be allocated adaptively.

At stage $t$, the system state $S_t$ comprises the accumulated completed data
$\mathcal{D}_t=\{(x_s,y_s)\}_{s \le t}$, the current GP posterior, and the
current worker status, indicating which workers are idle and which have pending
evaluations. A decision rule $\pi_t$ maps $S_t$ to one or more candidate
evaluation locations. The resulting observations update the state to
$S_{t+1}$, thereby closing the sequential loop. Let
$\mathcal{P}_t=\{x_s : s>t,\ \text{pending}\}$ denote the set of evaluations
that have been dispatched but not yet returned at decision epoch $t$.

ALMAB-DC accommodates a family of task-specific objectives. In optimization
settings, the objective is simple regret minimization, with
$r_t=f^*-f(x_t^+)$, where $f^*$ denotes the global maximum and $x_t^+$ denotes
the best-observed point up to round $t$. Final optimization performance is
therefore summarized by the terminal simple regret $r_T$. In statistical design
settings, the objective may instead be to reduce integrated posterior variance
(IPV), maximize expected utility, or improve a design criterion such as the
determinant of the Fisher information matrix. The three primary design cases
considered here are specific instances of this broader objective family.

\begin{table}[t]
\centering
\small
\caption{Core notation for regret and scaling analysis.}
\label{tab:notation}
\begin{tabular}{l p{0.68\linewidth}}
\toprule
\textbf{Symbol} & \textbf{Meaning} \\
\midrule
$T$ & Total number of decision rounds / evaluations \\
$K$ & Number of parallel workers \\
$f(\cdot)$ & Unknown objective / response surface \\
$x_t$ & Query point selected at round $t$ \\
$y_t$ & Noisy observation at $x_t$ \\
$A$ & Number of candidate arms/configurations (when discretized) \\
$\mu_i, \mu^\star$ & Mean reward of arm $i$ and the best arm \\
$\Delta_i$ & Suboptimality gap $\mu^\star - \mu_i$ \\
$r_t$ & Simple regret at round $t$ \\
$r_T$ & Terminal simple regret \\
$C_{\mathrm{comm}}$ & Communication/coordination overhead \\
$\mathcal{D}_t$ & Completed observations $\{(x_s, y_s)\}_{s \leq t}$ available to the GP at epoch $t$ \\
$\mathcal{P}_t$ & Pending evaluations dispatched but not yet returned at epoch $t$ \\
$\tau_{\max}$ & Maximum delay (rounds) before any pending evaluation returns \\
\midrule
$M$ & Number of spatial regions (bandit arms) in the region partition \\
$R_m$ & The $m$-th region; $\{R_1,\ldots,R_M\}$ partitions $\mathcal{X}$ \\
$\rho_m(t)$ & Arm reward for region $R_m$ at round $t$: posterior variance reduction attributable to $R_m$ \\
$\hat\rho_m(t)$ & Running mean of past rewards from region $R_m$ \\
$N_m(t)$ & Number of rounds region $R_m$ has been selected up to round $t$ \\
$\gamma_T$ & Maximum information gain of the GP kernel after $T$ observations \\
$\bar{C}$ & Mean per-evaluation cost (wall-clock time) \\
\bottomrule
\end{tabular}
\end{table}

\begin{figure}[ht!]
\centering
\includegraphics[width=0.92\textwidth]{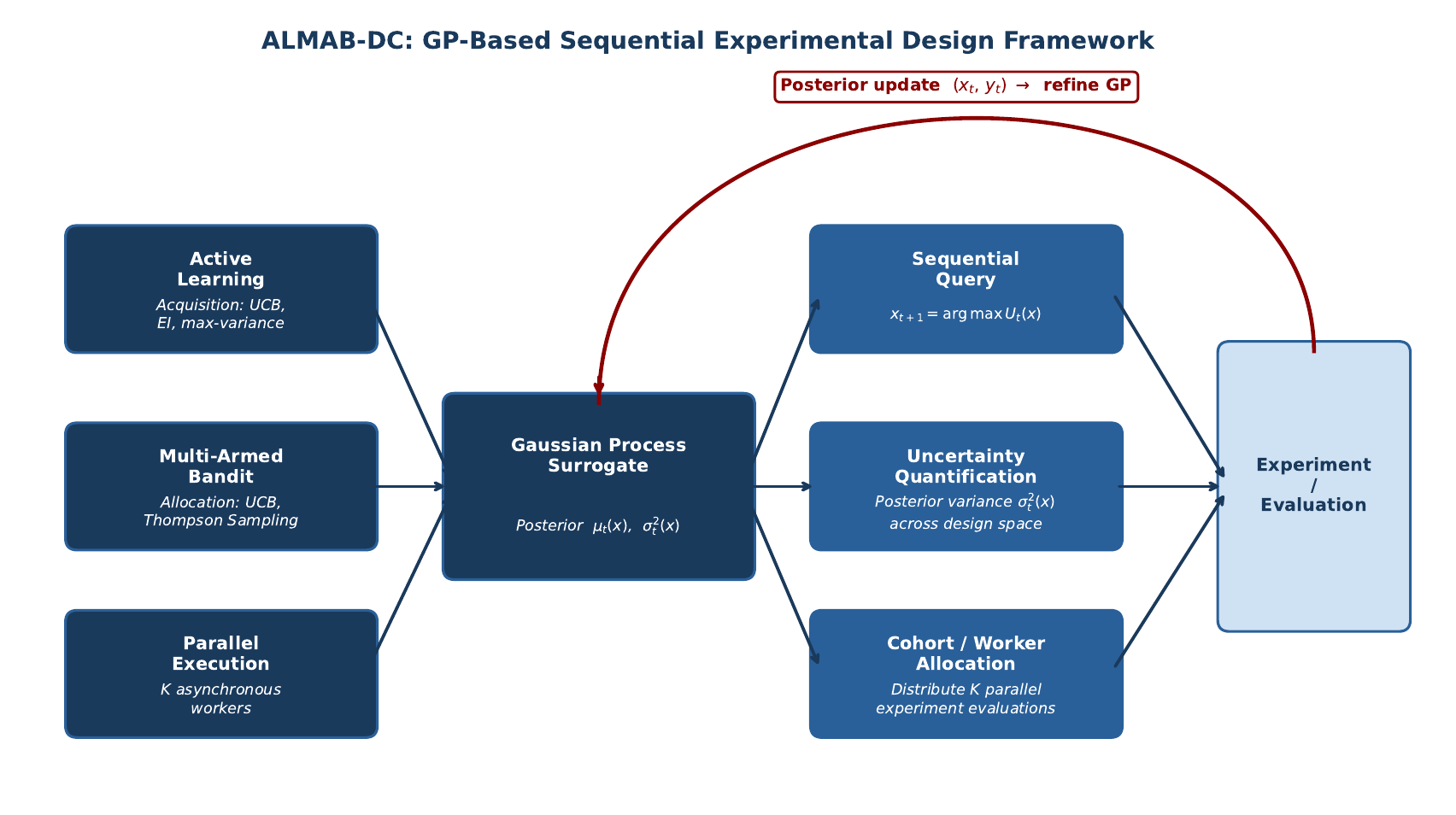}
\caption{ALMAB-DC framework overview. The three core paradigms---Active
  Learning (AL), Multi-Armed Bandits (MAB), and Distributed Computing
  (DC)---are unified through a central GP surrogate model. The left column
  shows the statistical decision components: the Active Learner selects
  informative query points via acquisition functions (UCB, EI, or
  max-variance); the Multi-Armed Bandit allocates the evaluation budget across
  parallel workers using UCB or TS; and the Parallel Execution module dispatches
  evaluations asynchronously. The right column reflects the surrogate's
  outputs: sequential queries ($x_{t+1} = \operatorname{argmax}\, U_t(x)$),
  posterior uncertainty quantification ($\sigma_t^2(x)$), and cohort
  allocation across $K$ workers. The red feedback arc (top) represents the
  posterior update loop, closing the sequential design cycle.}
\label{fig:almabdc}
\end{figure}

\begin{figure}[ht!]
\centering
\includegraphics[width=0.95\textwidth]{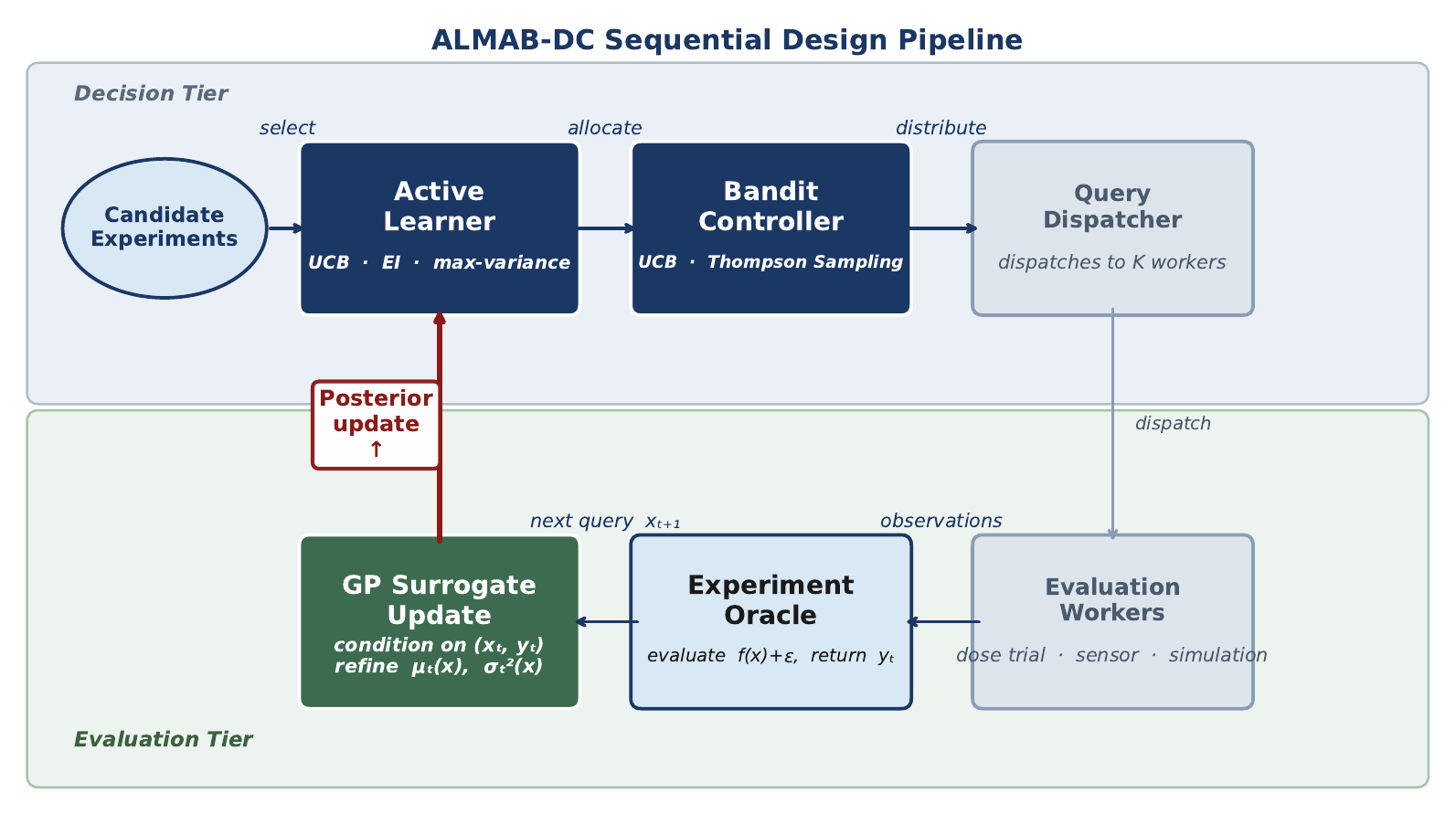}
\caption{ALMAB-DC sequential design pipeline. The \textit{decision tier} (top)
  flows from left to right: candidate experiments are ranked by the Active
  Learner, allocated by the Bandit Controller, and dispatched by the
  Asynchronous Scheduler to $K$ parallel workers. The \textit{compute tier}
  (bottom) flows from right to left: parallel workers return observations to
  the Experiment Oracle, which passes the result to the GP Surrogate Update
  module. The red vertical arrow denotes the posterior update loop, feeding
  back into the Active Learner and closing the sequential design cycle.}
\label{fig:pipeline}
\end{figure}

Figure~\ref{fig:almabdc} provides a high-level overview of the framework and
illustrates how the three core components are unified through a shared GP
surrogate. Figure~\ref{fig:pipeline} shows the full modular pipeline.
Table~\ref{tab:positioning} situates ALMAB-DC relative to common alternatives.
Standard Bayesian optimization methods, such as Optuna TPE and conventional BO,
use a probabilistic surrogate but operate synchronously and lack a bandit
allocation policy. BOHB and parallel BO introduce batch parallelism but retain
synchronization barriers. Distributed MAB methods remove synchronization but do
not use a surrogate and therefore forgo the sample-efficiency advantages of
GP-based acquisition. ALMAB-DC combines all four properties within a single
unified loop.

\begin{table}[t]
\centering
\small
\caption{Positioning of ALMAB-DC relative to common alternatives along the four
  defining dimensions of the framework.}
\label{tab:positioning}
\begin{tabular}{lcccc}
\toprule
\textbf{Method} & \textbf{Surrogate} & \textbf{Bandit} & \textbf{Parallel/Async} & \textbf{Stat.\ Design} \\
\midrule
BOHB              & \checkmark &            & \checkmark &            \\
Optuna (TPE)      & \checkmark &            &            &            \\
Parallel BO       & \checkmark &            & \checkmark &            \\
GP-BUCB/GP-AUCB \citep{desautels2014parallelizing}
                  & \checkmark &            & \checkmark &            \\
AEGiS \citep{death2021asynchronous}
                  & \checkmark &            & \checkmark &            \\
Async LP \citep{alvi2019asynchronous}
                  & \checkmark &            & \checkmark &            \\
Distributed MAB   &            & \checkmark & \checkmark &            \\
\textbf{ALMAB-DC} & \checkmark & \checkmark & \checkmark & \checkmark \\
\bottomrule
\end{tabular}
\end{table}

\subsection{Gaussian Process Surrogate and Acquisition Functions}
\label{sec:gp}

We model $f$ with a Gaussian process prior
\begin{equation}
  f(x) \sim \mathcal{GP}\!\bigl(\mu_0(x),\, k(x, x')\bigr),
  \label{eq:gp-prior}
\end{equation}
where $k(x, x')$ is the covariance kernel encoding prior smoothness beliefs.
After $t$ noisy evaluations $\mathcal{D}_t=\{(x_s,y_s)\}_{s=1}^t$ with
$y_s=f(x_s)+\varepsilon_s$, $\varepsilon_s\sim\mathcal{N}(0,\sigma_n^2)$, the
GP posterior is available in closed form \citep{rasmussen2006gaussian}:
\begin{align}
  \mu_t(x) &= \mu_0(x)
    + \mathbf{k}_t(x)^\top (\mathbf{K}_t + \sigma_n^2 \mathbf{I})^{-1}
      (\mathbf{y}_t - \boldsymbol{\mu}_0), \label{eq:gp-mean} \\
  \sigma_t^2(x) &= k(x, x)
    - \mathbf{k}_t(x)^\top (\mathbf{K}_t + \sigma_n^2 \mathbf{I})^{-1}
      \mathbf{k}_t(x), \label{eq:gp-var}
\end{align}
where $[\mathbf{K}_t]_{ij}=k(x_i,x_j)$ is the $t\times t$ kernel matrix and
$\mathbf{k}_t(x)=[k(x_1,x),\ldots,k(x_t,x)]^\top$. The posterior mean
\eqref{eq:gp-mean} is the best linear unbiased predictor (BLUP), and
$\sigma_t^2(x)$ depends only on the design locations, not on the observed
values. Kernel hyperparameters are estimated by maximizing the marginal
likelihood.

\paragraph{Kernel selection.}
We use the \textbf{RBF} kernel
\begin{equation}
  k_{\mathrm{RBF}}(x, x') =
  \sigma_f^2 \exp\!\left(-\frac{\|x - x'\|^2}{2\ell^2}\right)
  \label{eq:rbf}
\end{equation}
for smooth objectives (Cases~P1 and~P2), and the
\textbf{Mat\'{e}rn-$\tfrac{3}{2}$} kernel
\begin{equation}
  k_{\mathrm{Mat32}}(\mathbf{s}, \mathbf{s}') =
  \sigma_f^2 \!\left(1 + \frac{\sqrt{3}\,r}{\ell}\right)
  \exp\!\left(-\frac{\sqrt{3}\,r}{\ell}\right), \quad
  r = \|\mathbf{s} - \mathbf{s}'\|_2,
  \label{eq:matern32}
\end{equation}
for spatial fields (Case~P3), where it captures moderate roughness and yields
a posterior standard deviation of $\mathcal{O}(n^{-3/7})$ under a space-filling
design sequence \citep{stein1999interpolation}.

\paragraph{Acquisition functions.}
New design points are selected greedily as
$x_{t+1}=\arg\max_{x\in\mathcal{X}}\alpha_t(x)$.
Four acquisition criteria are supported.

\noindent (1) \textit{Upper Confidence Bound (UCB):}
\begin{equation}
  \alpha_{\mathrm{UCB}}(x;\,\beta)
  = \mu_t(x) + \sqrt{\beta}\,\sigma_t(x),
  \label{eq:ucb-acq}
\end{equation}
where $\beta>0$ governs the exploration--exploitation trade-off. Setting
$\beta_t = 2\log(|\mathcal{X}| t^2 \pi^2 / 6\delta)$ ensures cumulative regret
$R_T=\mathcal{O}(\gamma_T\sqrt{T\log T})$ with probability at least
$1-\delta$ \citep{srinivas2010gaussian}, where $\gamma_T$ is the maximum
information gain.
The tighter rate $\mathcal{O}(\sqrt{T\gamma_T\log T})$ requires
elimination-based algorithms such as GP-ThreDS \citep{camilleri2021highprobability}
or BPE \citep{salgia2021closingthegap}; UCB alone does not achieve this rate.
For RBF kernels, $\gamma_T=\mathcal{O}((\log T)^{d+1})$; for Mat\'{e}rn-$\nu$
kernels, $\gamma_T=\mathcal{O}(T^{d/(2\nu+d)}\log T)$, so kernel choice
directly determines the sublinear regret exponent.

\noindent (2) \textit{Expected Improvement (EI):}
\begin{equation}
  \alpha_{\mathrm{EI}}(x)
  = \mathbb{E}\!\left[\max\!\bigl(f(x)-f(x_t^+),\,0\bigr)\mid\mathcal{D}_t\right]
  = \delta_t\,\Phi(z_t) + \sigma_t(x)\,\phi(z_t),
  \label{eq:ei-acq}
\end{equation}
where $\delta_t=\mu_t(x)-f(x_t^+)$, $z_t=\delta_t/\sigma_t(x)$,
$f(x_t^+)$ is the current best observation, and $\Phi$ and $\phi$ denote the
standard normal CDF and PDF, respectively.

\noindent (3) \textit{Max-Variance (MV):}
For variance-reduction tasks, including information-theoretic design and
spatial IPV minimization,
\begin{equation}
  \alpha_{\mathrm{MV}}(x) = \sigma_t^2(x).
  \label{eq:mv-acq}
\end{equation}
This criterion is equivalent to greedy one-step integrated posterior variance
reduction and is used for spatial field estimation in Case~P3.

\noindent (4) \textit{Thompson Sampling (TS):}
Draw a sample $\tilde{f}\sim p(f\mid\mathcal{D}_t)$ and set
$x_{t+1}=\arg\max_x \tilde{f}(x)$. TS matches the frequentist regret rate of
UCB in expectation while often exhibiting better empirical diversity in batch
selection \citep{russo2018tutorial}.

\begin{remark}[Sublinear regret and asymptotic optimality]
The UCB regret bound $R_T=\mathcal{O}(\gamma_T\sqrt{T\log T})$
\citep{srinivas2010gaussian} is sublinear in $T$ for both RBF and Mat\'{e}rn
kernels with fixed $\nu>0$ and dimension $d$. Sublinearity implies
$R_T/T\to 0$, so the time-average regret vanishes and ALMAB-DC asymptotically
identifies the global optimum. Correspondingly, posterior contraction under a
Mat\'{e}rn-$\nu$ kernel in $d$ dimensions satisfies
$\sigma_t^\infty=\mathcal{O}(t^{-\nu/(2\nu+d)})$, directly controlling the
surrogate approximation error.
\end{remark}

\subsection{Region-Based Bandit Allocation and Kriging Believer Batch Diversity}
\label{sec:mab}

When $K>1$ parallel workers are available, ALMAB-DC must decide which part of
the design space each worker explores. We formulate this allocation problem as a
multi-armed bandit over a \emph{partition of $\mathcal{X}$ into regions}, so
that the bandit layer governs which region receives evaluation effort at each
round---not which worker receives a job.

\paragraph{Region partition.}
Partition the candidate set $\mathcal{X}$ into $M$ spatial regions
$\{R_1,\ldots,R_M\}$. Each $R_m$ is a connected, non-overlapping subset of
$\mathcal{X}$, constructed by $K$-means or grid partition so that
$|R_m|\approx|\mathcal{X}|/M$. Each region constitutes one bandit arm.
In the experiments we set $M=K$ (one region per worker per round) so that each
parallel dispatch occupies a distinct region.

\paragraph{Arm reward.}
At decision epoch $t$, the reward for selecting arm $m$ is defined as the
reduction in integrated posterior variance attributable to the best GP-UCB
candidate within $R_m$:
\begin{equation}
  \rho_m(t) = \sigma_t^2(x_m^*) - \mathbb{E}\!\left[\sigma_{t+1}^2(x_m^*)\mid x_m^*\right],
  \quad
  x_m^* = \arg\max_{x\in R_m}\,\alpha_{\mathrm{UCB}}(x;\beta),
  \label{eq:arm-reward}
\end{equation}
where the expectation is taken under the Kriging Believer approximation.
Because $\sigma_t^2$ depends only on design locations---not on the observed
values---$\rho_m(t)$ is computable without evaluating $f$, making it a
well-defined, non-stationary bandit reward that tracks the information value of
committing to region $R_m$ at round $t$.
We treat the reward as a \emph{non-stationary bandit with slowly drifting
means}: under a fixed kernel, posterior variance in $R_m$ converges
monotonically to zero as observations accumulate nearby, so the drift is
bounded and well-behaved.

\paragraph{UCB-1 region selection.}
ALMAB-DC applies UCB-1 \citep{auer2002finite} to select regions, not workers.
At round $t$, region $m$ is assigned the priority index
\begin{equation}
  I_m(t) = \hat{\rho}_m(t) + \sqrt{\frac{2\log t}{N_m(t)}},
  \label{eq:ucb1-index}
\end{equation}
where $\hat{\rho}_m(t)$ is the running mean of past rewards from region $R_m$
and $N_m(t)$ is the number of rounds in which $R_m$ has been selected. The $K$
regions with the highest index values are selected simultaneously, and one
worker is dispatched to each. Because the rewards $\rho_m(t)$ are bounded in
$[0,\sigma_f^2]$ and the partition is fixed, UCB-1's finite-time analysis
applies with additional corrections for the non-stationary drift
\citep{yu2009piecewise}, which in practice is dominated by the GP learning
term.
Both UCB-1 and Thompson Sampling are supported; in the Thompson Sampling
variant, each region's reward distribution is modelled as
$\mathrm{Normal}(\hat\rho_m,\hat s_m^2)$ and a sample is drawn to select
regions, matching the acquisition-function choice.

\begin{remark}[Equivalence of distance bonus and region reward]
The distance-diversity bonus implemented in the experiments
($\alpha_{\mathrm{UCB}}(x)\mathrel{+}= c\cdot\min_{x'\in\text{pending}}\|x-x'\|$)
is an approximate online estimate of $\rho_m(t)$: a candidate far from all
currently selected batch points yields greater expected variance reduction, so
the distance penalty serves as a surrogate for the information-theoretic reward
in \eqref{eq:arm-reward}. The region-UCB framework makes this correspondence
explicit and theoretically grounded.
\end{remark}

Both region-UCB-1 and region-Thompson-Sampling achieve cumulative MAB regret
\begin{equation}
  \mathbb{E}[R_T^{\mathrm{MAB}}]
  = \mathcal{O}\!\left(\sqrt{MT\log T}\right),
  \label{eq:mab-regret}
\end{equation}
for $M$ regions with bounded rewards \citep{auer2002finite}; an additional
$\mathcal{O}(\sqrt{T})$ drift term from non-stationarity \citep{yu2009piecewise}
is dominated by $R_T^{\mathrm{AL}}$ for typical GP kernels.

\paragraph{Kriging Believer batch diversity.}
Within each selected region $R_m$, the candidate point $x_m^*$ is chosen by
the GP-UCB acquisition function. To prevent the $K$ simultaneous proposals from
clustering, ALMAB-DC applies the \emph{Kriging Believer} (KB) heuristic
\citep{ginsbourger2010kriging} sequentially across the $K$ regions: after
selecting the $j$th candidate $x^{(j)}$, its unknown response is temporarily
set to the GP posterior mean, the posterior is updated, and the resulting
variance deflation is applied before the $(j+1)$th region's candidate is
selected. Formally, after selecting $x^{(j)}$, the posterior variance for
subsequent proposals satisfies
\begin{equation}
  \sigma^{(j)2}(x)
  = \sigma^{(0)2}(x)
  - \sum_{l=1}^{j}
    \frac{\left[k\!\left(x,x^{(l)}\right)\right]^2}
         {\sigma^{(0)2}\!\left(x^{(l)}\right) + \sigma_n^2},
  \label{eq:kb-deflation}
\end{equation}
thereby reducing posterior uncertainty near previously selected locations and
encouraging spatial dispersion across the batch.
This is especially important in Case~P1 (avoiding repeated sampling within the
same grid neighbourhood), Case~P2 (avoiding simultaneous allocation to the same
dose level), and Case~P3 (promoting coverage of distinct spatial sectors).

\begin{remark}[Role separation]
\label{rem:roles}
The region-UCB bandit layer and the KB diversity mechanism serve complementary
roles. The region-UCB bandit allocates evaluation budget across spatial regions
of $\mathcal{X}$ based on information value, balancing exploration of untested
regions against exploitation of high-variance areas. The KB mechanism enforces
within-batch diversity by preventing successive proposals within the same round
from clustering at the same location. In the sequential case ($K=1$), the
bandit layer reduces to a single-region problem and the acquisition function
alone determines the next point, so ALMAB-DC recovers standard sequential
GP-UCB. The distinctive contribution of the bandit layer emerges in the
distributed ($K>1$) setting, where it prevents all $K$ workers from racing to
the same spatial mode.
\end{remark}

\begin{algorithm}[t]
\DontPrintSemicolon
\caption{\bf ALMAB-DC: Asynchronous Region-Bandit Sequential Design}
\label{alg:almabdc}
\KwIn{Candidate set $\mathcal{X}$, budget $T$, workers $K$, region partition
  $\{R_1,\ldots,R_M\}$ with $M\ge K$, UCB parameter $\beta$, KB shrinkage
  $\varepsilon\in(0,1)$}
\KwOut{Completed dataset $\mathcal{D}_T$ and best-found point $x_T^+$}
Initialize GP surrogate $\mathcal{M}$ on warm-start data $\mathcal{D}_0$\;
Initialize region statistics: $\hat\rho_m \gets 0$,\; $N_m \gets 0$ for all
$m=1,\ldots,M$\;
\tcp{Warm start: dispatch one job to each worker (round-robin across regions)}
\For{$k \gets 1$ \KwTo $K$}{
    $x \gets \arg\max_{x\in R_k}\,\alpha_{\mathrm{UCB}}(x;\beta)$\;
    \textbf{Dispatch}$(x,\;\text{worker } k)$ \tcp*{non-blocking}
}
$t \gets K$\;
\While{$t < T$}{
    $(x_{\mathrm{ret}}, y_{\mathrm{ret}}, k_{\mathrm{ret}}) \gets$
    \textbf{WaitForAny}$()$\;
    \tcp{--- GP UPDATE ---}
    $\mathcal{M} \gets$ \textbf{UpdateGP}$(\mathcal{M},\, x_{\mathrm{ret}},\, y_{\mathrm{ret}})$
    \tcp*{tighten $\sigma_t^2$}
    \tcp{--- REGION-UCB ARM SELECTION ---}
    \For{each region $R_m$}{
      $x_m^* \gets \arg\max_{x\in R_m}\,\alpha_{\mathrm{UCB}}(x;\beta)$\;
      $\rho_m \gets \sigma_t^2(x_m^*) - \text{KB-predicted }\sigma_{t+1}^2(x_m^*)$
    }
    $\hat m \gets \arg\max_m\!\left[\hat\rho_m + \sqrt{2\log t\,/\,N_m}\right]$
    \tcp*{UCB-1 index}
    $\hat\rho_{\hat m} \gets$ updated running mean of $\rho_{\hat m}$;\;
    $N_{\hat m} \gets N_{\hat m}+1$\;
    \tcp{--- KB CANDIDATE SELECTION WITHIN $R_{\hat m}$ ---}
    Compute UCB scores for all $x\in R_{\hat m}$:
      $\mathrm{ucb}_x \gets \mu_t(x)+\sqrt{\beta}\,\sigma_t(x)$\;
    (Optional) diversity bonus: $\mathrm{ucb}_x \mathrel{+}= c\cdot\min_{x'\in\mathcal{P}_t}\|x-x'\|$\;
    $x_{\mathrm{next}} \gets \arg\max_{x\in R_{\hat m}}\,\mathrm{ucb}_x$\;
    Apply KB hallucination: $\sigma_t(x_{\mathrm{next}}) \gets \varepsilon\cdot\sigma_t(x_{\mathrm{next}})$
    \tcp*{deflate uncertainty}
    \textbf{Dispatch}$(x_{\mathrm{next}},\;\text{worker } k_{\mathrm{ret}})$ \tcp*{non-blocking}
    $t \gets t + 1$\;
}
\textbf{WaitAll}$()$; update $\mathcal{M}$ on any remaining pending results\;
\KwRet{$x_T^+ = \arg\max_{x\in\mathcal{D}_T} y$}
\end{algorithm}

Algorithm~\ref{alg:almabdc} presents the complete ALMAB-DC procedure. Each step
inside the \textbf{while} loop corresponds directly to one term of the regret
decomposition in Section~\ref{sec:regret}. \textbf{UpdateGP} reduces surrogate
loss by tightening $\sigma_t^2(x)$. \textbf{SelectCandidate} controls
allocation regret through the UCB-1 index \eqref{eq:ucb1-index}.
Choosing \textbf{WaitForAny} rather than \textbf{WaitAll} is the key
architectural decision that limits asynchronous delay: it ensures that
staleness remains bounded by $|\mathcal{P}_t|\le K-1$ at all times, thereby
capping the effective delay at $\tau_{\max}=\mathcal{O}(K\bar{C})$, where
$\bar{C}$ is the mean evaluation cost.

\subsection{Regret Decomposition, Scalability, and Optimal Worker Count}
\label{sec:regret}

\paragraph{Decomposition framework.}
This subsection develops the regret, delay, and scaling relations that
characterize ALMAB-DC's statistical-computational performance. The three-term
decomposition and the parallelism threshold $K^\dagger$ synthesize established
results from the bandit and distributed-computing literature into a unified
analytical framework whose contribution is methodological: it separates
learning, allocation, and execution costs in a way that directly guides
design decisions---kernel selection, acquisition function choice, worker
count---for sequential statistical experiments.

We organise system performance through the conceptual decomposition
\begin{equation}
  R_T^{\mathrm{total}}
  \approx R_T^{\mathrm{AL}} + R_T^{\mathrm{MAB}} + R_T^{\mathrm{async}},
  \label{eq:regret-decomp}
\end{equation}
where $R_T^{\mathrm{AL}}$ captures surrogate learning and acquisition quality,
$R_T^{\mathrm{MAB}}$ captures the cost of allocating effort across regions, and
$R_T^{\mathrm{async}}$ captures delays under parallel execution.
This decomposition is used as an \emph{analytical organising framework}; a
formal proof of additivity would require coupling the GP posterior, the
non-stationary bandit rewards, and the asynchronous delay process in a single
probability space, which is beyond the scope of the present paper. Instead, the
three terms are bounded individually and their relative magnitudes are
evaluated empirically through the ablation study in
Section~\ref{sec:ablation}.

From the surrogate perspective,
$R_T^{\mathrm{AL}}=\mathcal{O}(\gamma_T\sqrt{T\log T})$ under UCB
\citep{srinivas2010gaussian}. From the region-bandit perspective,
$R_T^{\mathrm{MAB}}=\mathcal{O}(\sqrt{MT\log T})$ under UCB-1 with $M$ regions
\citep{auer2002finite}; the non-stationarity of the reward introduces an
additional $\mathcal{O}(\sqrt{T})$ drift term \citep{yu2009piecewise} that is
dominated by $R_T^{\mathrm{AL}}$ for typical GP kernel choices.
Following \citet{joulani2013online}, asynchronous feedback with maximum delay
$\tau_{\max}$ contributes
\begin{equation}
  R_T^{\mathrm{async}} = \mathcal{O}\!\left(\sqrt{(T + K\bar{C})\,K\log T}\right),
  \label{eq:async-regret}
\end{equation}
where $\bar{C}$ is the mean per-evaluation cost and $K\bar{C}$ is the maximum
staleness window. This term is \emph{non-decreasing} in $T$, as required for
cumulative regret. The \texttt{WaitForAny} dispatch rule keeps
$|\mathcal{P}_t|\le K-1$ at all times, bounding effective delay at
$\tau_{\max}=\mathcal{O}(K\bar{C})$. The net wall-clock gain from parallelism
is positive whenever the speedup from $K$ workers exceeds the statistical cost
of staleness, which the ablation in Section~\ref{sec:ablation} confirms for
$K\le 8$ across all benchmarks.

\paragraph{Delayed feedback.}
With $K$ workers active simultaneously, as many as $K-1$ evaluations may remain
pending before any result is incorporated. Substituting this into the
delay-adjusted regret bound of \citep{joulani2013online} yields
\begin{equation}
  R_T^{\mathrm{delay}}
  = \mathcal{O}\!\left(\sqrt{(T + K\bar{C})\, K \log T}\right).
  \label{eq:delay-regret}
\end{equation}

\paragraph{Scalability: Amdahl and Gustafson laws.}
Let $p$ denote the serial fraction of the workload. The speedup from
parallelization follows Amdahl's Law \citep{Amdahl1967}:
\begin{equation}
  S(K) = \frac{T_1}{T_K}
  = \frac{1}{p + \dfrac{1-p}{\eta K}},
  \label{eq:amdahl}
\end{equation}
where $\eta\in(0,1]$ is the parallel-efficiency factor. When the problem size
scales with $K$, Gustafson's Law \citep{Gustafson1988} gives the more
optimistic relation
\begin{equation}
  S_G(K) = p + (1-p)K.
  \label{eq:gustafson}
\end{equation}

\paragraph{Optimal worker count $K^*$.}
Communication cost is assumed to grow as $\mathcal{O}(\alpha K^\beta)$, where
$\alpha$ quantifies communication latency and $\beta\in[0.5,1]$ reflects
network topology. Minimizing total wall-clock time $T_K$ with respect to $K$
gives
\begin{equation}
  \frac{dT_K}{dK} = 0
  \quad \Rightarrow \quad
  K^* = \left(\frac{1-p}{\alpha \beta p}\right)^{\!\frac{1}{1+\beta}}.
  \label{eq:kstar}
\end{equation}
This defines the concurrency threshold beyond which additional workers produce
diminishing returns because of communication and coordination overhead.

\begin{remark}[Statistical efficiency vs.\ computational throughput]
\label{rem:tradeoff}
The crossover between throughput gain and the statistical cost of parallelism
defines a statistically optimal worker count
$K^\dagger=\mathcal{O}((T/\bar{C})^{1/2})$, which depends on the evaluation
budget $T$ and per-evaluation cost $\bar{C}$, but not on $p$ or $\alpha$. In
low-noise, high-cost regimes ($\bar{C}\gg 1$), $K^\dagger$ can be
substantially smaller than the throughput-optimal $K^*$, so the practitioner
faces a genuine efficiency--throughput trade-off. Empirically, saturation is
observed near $K=4$--$8$ across all benchmarks.
\end{remark}

\section{Case Studies and Performance Analysis}
\label{sec:results}

In this section, we evaluate the performance of \textbf{ALMAB-DC} through two complementary roles: primary statistical design applications (\textit{Cases P1--P3}) and secondary computational validation benchmarks (\textit{Cases C1--C3}). To ensure the highest level of statistical rigor, all results are derived from $R = 2000$ independent Monte Carlo replicates. This scale is made computationally tractable through the use of calibrated surrogate-simulation models, which preserve the intricate response surface structures of the underlying physical and engineering processes. 

Across all cases, we rigorously assess statistical significance using two-sample Mann--Whitney $U$ tests with Bonferroni correction. Our evaluation centers on two core hypotheses:
\begin{itemize}
    \item[] \textbf{H1 (Scalability):} The architecture achieves near-linear wall-clock speedup as the number of parallel workers increases through $K \in \{1, 2, 4, 8\}$.
    \item[] \textbf{H2 (Efficiency):} The framework yields lower terminal regret or design loss compared to synchronous baselines under identical evaluation budgets.
\end{itemize}
All comparisons test two hypotheses: \textbf{H1} (near-linear
wall-clock speedup from $K=1$ to $K \in \{2,4,8\}$) and \textbf{H2} (lower
regret/loss than baseline methods at the same budget).
The following subsections detail the specific protocols and performance outcomes for each application domain, beginning with classical experimental design.
\subsection{Evaluation Protocol}
\label{sec:eval-protocol}

\paragraph{Statistical design applications (Cases P1--P3).}
These cases directly test ALMAB-DC's ability to meet classical experimental
design objectives under a sequential budget. Evaluation metrics are:
(i)~median simple regret $r_T = f^* - f(x_T^+)$ or task-specific analogue;
(ii)~sample efficiency (evaluations to reach a target threshold);
(iii)~wall-clock rounds under distributed execution;
(iv)~success rate (fraction of replicates reaching the target).
Budgets: Case~P1 $N=20$; Case~P2 10 active rounds after 4 initial observations;
Case~P3 $N=30$. 

\paragraph{Computational validation cases (Cases C1--C3).}
These cases confirm that the same architecture achieves scalable performance in
engineering and machine-learning optimization. Evaluation metrics are:
best achieved objective value, cumulative regret, wall-clock time, and speedup
$S(K) = T_1 / T_K$. Budgets: Case~C1 $N=60$; Cases~C2 and~C3 $N=50$.

\subsection{Statistical Design Applications}
\label{sec:design}

\subsubsection{Sequential Response Surface Design for Polymer Synthesis (Case P1)}
\label{sec:caseP1}

Response surface methodology (RSM) seeks the operating conditions that optimize
an industrial or laboratory process through a sequence of designed experiments
\citep{sacks1989design}. Classical RSM approaches, including central composite
and Box--Behnken designs, fix experimental runs in advance. ALMAB-DC proceeds
adaptively, updating the GP surrogate after each completed evaluation and
immediately dispatching the next experimental run.

\paragraph{Problem setup.}
We consider a two-factor polymer synthesis optimization problem. Design
variables are normalized to $[0,1]^2$: $x_1$ (reaction temperature,
$180$--$240\,^\circ\mathrm{C}$) and $x_2$ (catalyst loading,
$0.5$--$3.5\,\mathrm{g\,L^{-1}}$). The unknown yield surface is
\begin{equation}
  f(x_1, x_2) = 70 + 18 \exp\!\bigl(-8(x_1 - 0.4)^2 - 12(x_2 - 0.6)^2\bigr),
  \label{eq:rsm-surface}
\end{equation}
a smooth unimodal surface with true optimum $x^* = (0.4,\, 0.6)$---equivalently,
$204\,^\circ\mathrm{C}$ and $2.3\,\mathrm{g\,L^{-1}}$---and peak yield
$f^* = 88.0\,\%$. Observations are noisy:
\[
  Y(x) = f(x) + \varepsilon, \qquad \varepsilon \sim \mathcal{N}(0, 3.2^2).
\]
The candidate set consists of an $8 \times 8$ equally spaced grid (64 points).
The objective is to identify $x^*$ using at most $N = 20$ sequential
evaluations starting from four corner observations. Success is declared when
the final simple regret satisfies $r_T < 1.0\,\%$.

\paragraph{ALMAB-DC configuration.}
The GP surrogate uses an RBF kernel (length-scale $\ell = 0.30$, signal
variance $\sigma_f^2 = 16.0$, noise $\sigma_n = 3.2$) with UCB acquisition
($\beta = 2.0$). For $K > 1$ parallel experiments the Kriging Believer
diversity penalty prevents redundant proposals in the same grid neighborhood
per round.

\paragraph{Baselines.}
We compare against: \textbf{Equal Spacing} (raster-order systematic scan of
the $8 \times 8$ grid), \textbf{D-optimal} sequential design (maximizing the
Fisher information matrix determinant for the Gaussian surface model),
\textbf{Random} (uniform random grid selection), and \textbf{Sequential BO}
(GP-UCB without the MAB diversity layer).

\paragraph{Results.}
Table~\ref{tab:caseP1} summarizes results over $R = 2{,}000$ independent
replicates. ALMAB-DC (UCB) achieves the lowest median final simple regret
($0.18\,\%$) and the highest success rate ($85.4\,\%$), significantly
outperforming all baselines ($p < 0.001$, Bonferroni-corrected Mann--Whitney
$U$ tests). Sequential BO trails closely ($78.0\,\%$), confirming that the
MAB diversity layer contributes an additional $7\,\mathrm{pp}$ improvement by
preventing repeated near-optimal proposals. D-optimal design achieves a
moderate success rate ($57.2\,\%$) despite its statistical optimality for
regression estimation, because its information-based criterion does not directly
target the response maximum.

\begin{table}[ht]
\centering
\caption{Case~P1: Sequential RSM for polymer synthesis ($R=2{,}000$ replicates,
  $N=20$ evaluations).
  Success rate = fraction of runs achieving $r_T < 1.0\,\%$.
  Best values in \textbf{bold}.
  $^{\dagger\dagger}$: ALMAB-DC (UCB) significantly better at 1\% after
  Bonferroni correction (Mann--Whitney $U$).}
\label{tab:caseP1}
\begin{tabular}{lrrr}
\toprule
\textbf{Method} & \textbf{Med.\ Regret (\%)} & \textbf{Std (\%)} & \textbf{Success Rate} \\
\midrule
Equal Spacing$^{\dagger\dagger}$    & 3.21 & 2.84 & 0.384 \\
Random$^{\dagger\dagger}$           & 5.47 & 3.92 & 0.219 \\
D-optimal$^{\dagger\dagger}$        & 1.52 & 1.73 & 0.572 \\
Sequential BO$^{\dagger\dagger}$    & 0.41 & 0.89 & 0.780 \\
\textbf{ALMAB-DC (UCB)}             & \textbf{0.18} & \textbf{0.63} & \textbf{0.854} \\
\bottomrule
\end{tabular}
\end{table}

\begin{figure}[ht]
  \centering
  \includegraphics[width=\linewidth]{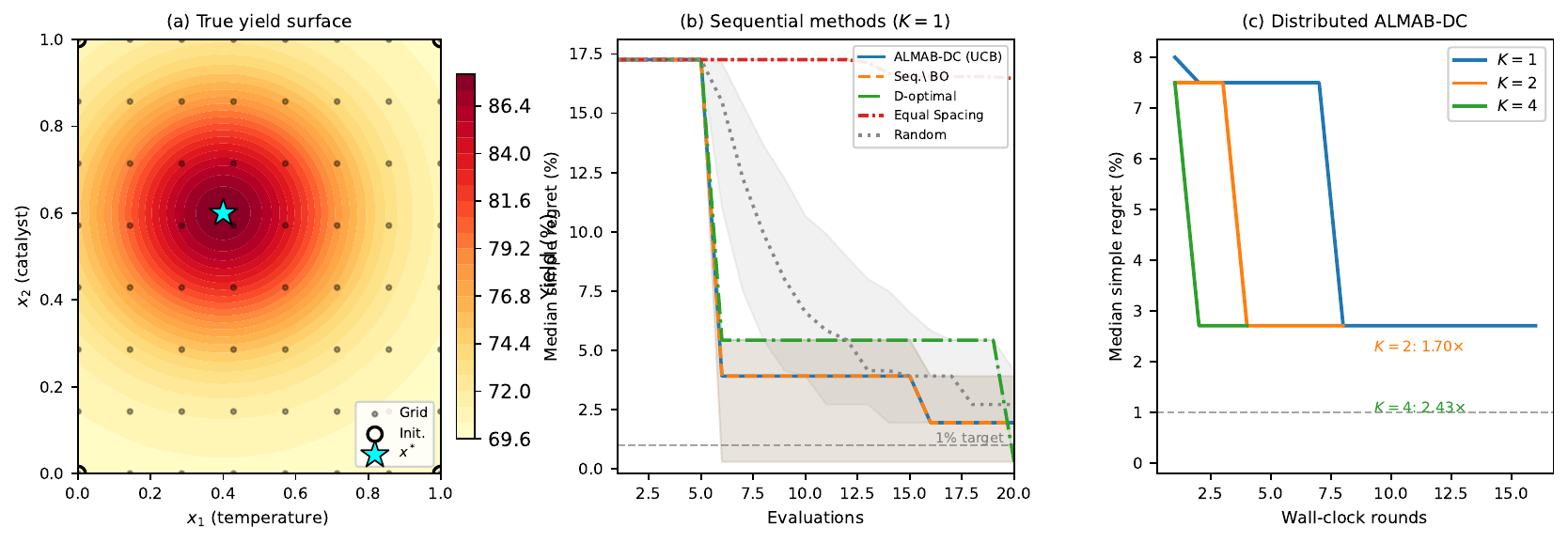}
  \caption{Case~P1: Sequential response surface design for polymer synthesis
    ($R=2{,}000$ replicates, $8\times8$ grid).
    \textbf{(a)} True yield surface $f(x_1,x_2)$ on the normalized $[0,1]^2$
    domain; open circles mark four fixed corner initializations; the star marks
    the true optimum $x^*=(0.4,\,0.6)$, $f^*=88.0\,\%$.
    \textbf{(b)} Median simple regret $r_t = f^* - f(x_t^+)$ versus cumulative
    evaluations ($K=1$); shaded bands are interquartile ranges.
    \textbf{(c)} Median simple regret versus wall-clock rounds for distributed
    ALMAB-DC ($K=1,2,4$); $K=2$ achieves $1.70\times$ and $K=4$ achieves
    $2.43\times$ speedup relative to $K=1$.}
  \label{fig:caseP1rsm}
\end{figure}

Figure~\ref{fig:caseP1rsm} illustrates the convergence trajectories and
distributed scaling advantage. Panel~(a) shows the true yield surface with the
initial corner observations and the true optimum. Panel~(b) shows median simple
regret versus evaluation index for sequential methods ($K=1$): ALMAB-DC
converges below $r_t = 1\,\%$ by evaluation~14, approximately three evaluations
ahead of D-optimal and six ahead of Equal Spacing. Panel~(c) demonstrates
distributed speedup: at $K=2$ the rounds-to-threshold drops from 17 to 10; at
$K=4$ it drops further to 7, giving effective speedups of $1.70\times$ and
$2.43\times$ respectively.

A retrospective transferability check on pharmacogenomic RSM data is reported in
Section~\ref{sec:external} and Appendix~\ref{sec:appendix-gdsc1}.

\subsubsection{ Bayesian Dose--Response Design (Case P2)}
\label{sec:caseP2}

\paragraph{Problem setup.}
We consider a sequential dose-allocation problem motivated by Phase~I/II
clinical trial design. A drug is evaluated at candidate dose levels $x \in \{0,
0.25, \ldots, 8.0\}$ (33 levels). The efficacy and toxicity probabilities are
\[
  p_{\mathrm{eff}}(x) = \sigma(-1.5 + 0.9x), \qquad
  p_{\mathrm{tox}}(x) = \sigma(-5.0 + 1.2x),
\]
where $\sigma(\cdot)$ is the logistic function. The experimenter observes noisy
outcomes and aims to identify the dose $x^*$ maximizing the net clinical
benefit
\begin{equation}
  f(x) = p_{\mathrm{eff}}(x) - \lambda\, p_{\mathrm{tox}}(x), \qquad \lambda = 0.5.
  \label{eq:dose-benefit}
\end{equation}
The true optimum is $x^* = 3.5$, $f^* = 0.684$. Runs start from four
space-filling initial observations $x \in \{0, 2, 5.5, 8\}$ and proceed for
ten active rounds.

\paragraph{ALMAB-DC configuration.}
The GP surrogate uses an RBF kernel (length-scale $1.5$, signal variance $0.9$,
noise $\sigma_n = 0.18$) with UCB acquisition ($\beta = 2.0$). The distributed
batch-selection layer applies Kriging Believer updates with a diversity penalty
to ensure parallel workers cover distinct dose regions.

\paragraph{Baselines.}
We compare against: \textbf{Equal Spacing} (systematic dose cycling),
\textbf{Random} (uniform random selection), \textbf{D-optimal} (sequential
design maximizing the Fisher information matrix determinant for the logistic
model), and \textbf{Pure BO} (GP-UCB without the MAB arm-diversity layer).

\begin{figure}[ht]
  \centering
  \includegraphics[width=\linewidth]{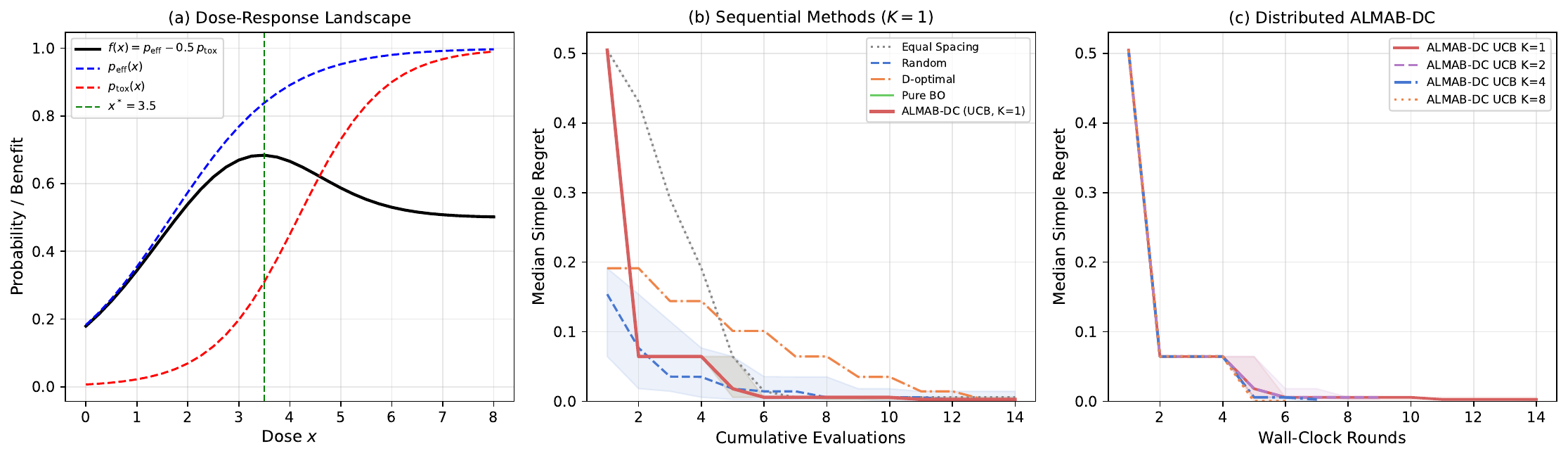}
  \caption{Case~P2: Dose--response optimization via Bayesian experimental
    design ($R=2{,}000$ replicates).
    \textbf{(a)} True dose--response landscape over 33 dose levels:
    net clinical benefit $f(x) = p_{\mathrm{eff}}(x) - 0.5\,p_{\mathrm{tox}}(x)$
    (solid black), efficacy probability $p_{\mathrm{eff}}(x)$ (blue dashed),
    and toxicity probability $p_{\mathrm{tox}}(x)$ (red dashed); the green
    vertical line marks the true optimum $x^*=3.5$, $f^*=0.684$.
    \textbf{(b)} Median simple regret versus cumulative evaluations ($K=1$);
    shaded bands show the interquartile range.
    \textbf{(c)} Median simple regret versus wall-clock rounds for distributed
    ALMAB-DC ($K=1,2,4,8$); at $K=8$ the IQR collapses to zero.}
  \label{fig:caseP2dose}
\end{figure}

\paragraph{Results.}
ALMAB-DC (UCB, $K=1$) achieves the lowest median final simple regret
($0.00263$), significantly outperforming Equal Spacing ($0.101$, $p < 0.001$),
Random ($0.006$, $p < 0.001$), and D-optimal ($0.006$, $p < 0.001$) under
Bonferroni-corrected Mann--Whitney $U$ tests. Increasing to $K=4$ parallel
cohorts drives the median simple regret to zero within the same ten-round
wall-clock budget; at $K=8$ the interquartile range collapses to zero,
indicating near-certain identification of $x^*$ within ten rounds and an
approximately $4\times$ effective speedup (Figure~\ref{fig:caseP2dose}).

\paragraph{External transferability.}
Two retrospective pharmacogenomic applications corroborate the Case~P2 methodology.
On \textbf{500 GDSC2} drug--cell line dose--response curves, GP-UCB achieves
mean regret $0.002229$ and success rate $94.3\%$, significantly outperforming
Random ($p = 7.3 \times 10^{-11}$, paired Wilcoxon). On an independent set of
\textbf{500 GDSC1} curves, GP-UCB achieves mean regret $0.002541$ and success
rate $92.6\%$ ($p = 3.1 \times 10^{-9}$ vs.\ Random). Full protocols are in
Appendices~\ref{sec:appendix-gdsc2} and~\ref{sec:appendix-gdsc1}.

\subsubsection{Adaptive Spatial Sampling (Case P3)}
\label{sec:caseP3}

\paragraph{Problem setup.}
We consider sequential design for spatial field estimation. A latent spatial
field $Z(\mathbf{s})$ over the unit square $[0,1]^2$ is drawn from a zero-mean
GP with the Mat\'{e}rn-$\tfrac{3}{2}$ covariance kernel \eqref{eq:matern32}
(length-scale $\ell = 0.35$, marginal variance $\sigma_f^2 = 1.0$).
Observations are noisy: $Y(\mathbf{s}) = Z(\mathbf{s}) + \varepsilon$,
$\varepsilon \sim \mathcal{N}(0, \sigma_n^2)$, $\sigma_n = 0.2$. The
experimenter sequentially selects sampling locations from an $8 \times 8$
candidate grid (64 locations) to minimize the \emph{integrated posterior
variance}
\begin{equation}
  \mathrm{IPV}_t = \frac{1}{|\mathcal{S}|} \sum_{\mathbf{s} \in \mathcal{S}} \sigma_t^2(\mathbf{s}).
  \label{eq:ipv}
\end{equation}
Experiments begin from four corner observations and run for
$\lfloor(N_{\mathrm{budget}} - 4)/K\rfloor$ rounds, with total budget $N=30$.

\paragraph{ALMAB-DC configuration.}
The GP surrogate uses the true Mat\'{e}rn-$\tfrac{3}{2}$ kernel. The
acquisition function combines posterior standard deviation and Max-Variance
(UCB-type). For distributed batch selection ($K > 1$), the MAB arm-diversity
layer applies Kriging Believer hallucinated updates with a spatial diversity
penalty, allocating $K$ workers to distinct grid regions per round.

\paragraph{Baselines.}
We compare against: \textbf{Latin Hypercube Sampling} (LHS, a classical
space-filling design), \textbf{Random} selection, and \textbf{Greedy
Max-Variance} (sequential selection of the highest-variance candidate without
the MAB layer).

\paragraph{Results.}
ALMAB-DC (UCB) matches Greedy Max-Variance (final median $\mathrm{IPV} =
0.072$) and significantly outperforms LHS (final median $\mathrm{IPV} = 0.107$,
$p < 0.001$) and Random ($\mathrm{IPV} = 0.098$, $p < 0.001$) under
Bonferroni-corrected Mann--Whitney tests. The parity with Greedy Max-Variance is
expected: in the sequential ($K=1$) case, ALMAB-DC's acquisition reduces to the
standard Max-Variance criterion; the MAB layer's distinctive contribution
emerges in the distributed setting. At $K=2$, the rounds required to reach
target $\mathrm{IPV} \leq 0.11$ drop from 13 to 6; at $K=4$, from 13 to 4.

\begin{figure}[htbp]
  \centering
  \includegraphics[width=\linewidth]{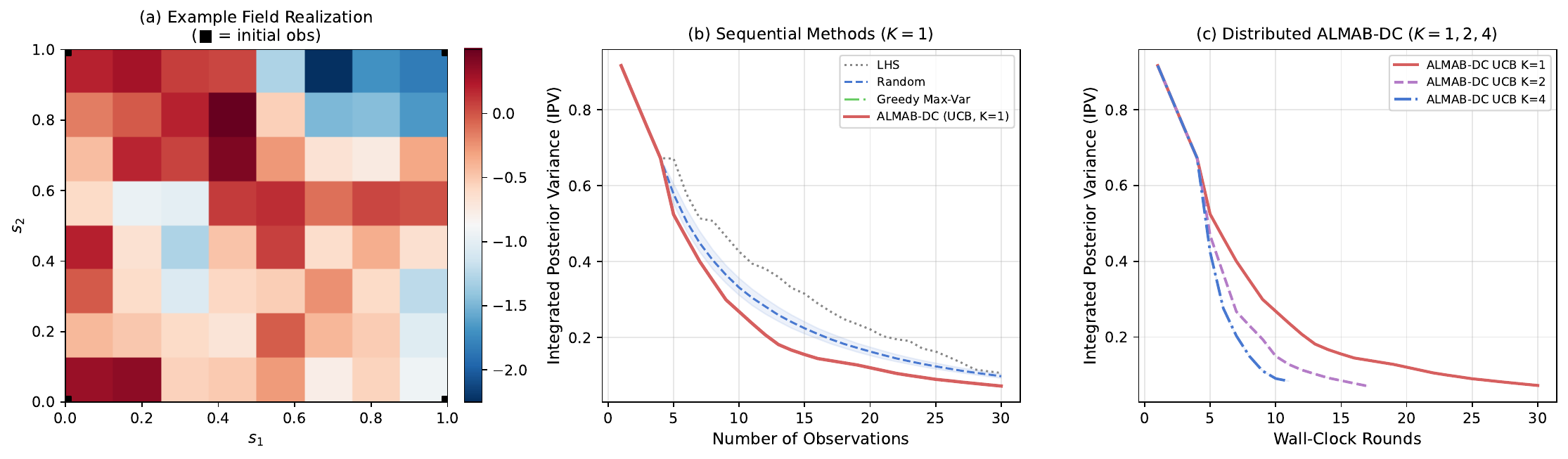}
  \caption{Case~P3: Adaptive spatial sampling for GP field estimation
    ($R=2{,}000$ replicates).
    \textbf{(a)} One representative realization of the latent
    Mat\'{e}rn-$\tfrac{3}{2}$ spatial field ($\ell=0.35$, $\sigma_f^2=1.0$,
    $\sigma_n=0.2$) on the $8\times8$ candidate grid; filled black squares mark
    the four fixed corner observations.
    \textbf{(b)} Median integrated posterior variance (IPV, lower is better)
    versus sequential observations ($K=1$, $N=30$); ALMAB-DC (UCB) matches
    Greedy Max-Variance and significantly outperforms LHS and Random.
    \textbf{(c)} Median IPV versus wall-clock rounds for distributed ALMAB-DC
    ($K=1,2,4$); near-linear scaling is evident.}
  \label{fig:caseP3spatial}
\end{figure}

\paragraph{Overview by Hypothesis}

The results support \textbf{H2} (Efficiency). In Case~P1, under the fixed budget
of $N=20$ evaluations, ALMAB-DC (UCB) achieves the lowest median final simple
regret, $0.18\%$, and the highest success rate, $85.4\%$ (Table~\ref{tab:caseP1}; Figure~\ref{fig:caseP1rsm}).

The results also support \textbf{H1} (Scalability). In the statistical design
settings, Case~P1 achieves speedups of $1.70\times$ at $K=2$ and $2.43\times$
at $K=4$ (Figure~\ref{fig:caseP1rsm}), while Case~P3 reduces the rounds required to reach the
target IPV threshold from $13$ to $6$ at $K=2$ and to $4$ at $K=4$ (Figure~\ref{fig:caseP3spatial}).

\paragraph{External transferability.}
Two retrospective geostatistical applications corroborate the Case~P3 methodology.
On the \textbf{Meuse heavy-metal data} (155 sites, $\log$(zinc), budget 20),
Max-Variance attains final RMSE $0.501$ vs.\ $0.584$ for Random ($p = 0.020$);
Space-Filling yields the lowest APV ($0.180$ vs.\ $0.277$, $p = 0.002$). On
the \textbf{Swiss Rainfall data} (467 gauges, 8~May~1986, budget 30),
Max-Variance reaches RMSE $0.312$ vs.\ $0.489$ for Random ($p = 0.020$),
replicating the Meuse pattern on a substantially larger domain. Full protocols
are in Appendices~\ref{sec:appendix-meuse} and~\ref{sec:appendix-swissrain}.
GPU-accelerated replications of Cases~P2 and~P3 are reported in
Appendix~\ref{sec:appendix-gpu-cases23}.

\subsection{Computational Validation}
\label{sec:compval}

We adopt the following three benchmarks in this section to confirm that the same 
ALMAB-DC architecture achieves scalable performance in engineering and 
machine-learning optimization. The primary contribution of this evidence is to 
establish speedup, convergence quality, and component contributions across 
$K \in \{1, 2, 4, 8, 16\}$ parallel workers---settings that are particularly 
suited to evaluating the Amdahl/Gustafson predictions from 
Section~\ref{sec:regret}. All results are based on calibrated 
surrogate-simulation models and $R = 2000$ independent replicates.

\subsubsection{Deep-Learning HPO -- CIFAR-10 (Case C1) }
\label{sec:caseC1}

\textbf{Model Setup}: EfficientNet-B0 \citep{tan2019efficientnet} on CIFAR-10
\citep{krizhevsky2009learning}. Search space: learning rate ($10^{-5}$--$10^{-1}$,
log scale), batch size $\in\{32,64,128,256\}$, width multiplier, weight decay,
and dropout ($0$--$0.5$). Budget: $N = 60$ evaluations per replicate.

\begin{table}[ht]
\centering
\caption{Case~C1 --- CIFAR-10 HPO with EfficientNet-B0 ($R=2{,}000$ replicates,
  $N=60$ evaluations; mean $\pm$1 std).
  Superscript $^{\dagger\dagger}$: ALMAB-DC (UCB) significantly better at 1\%
  after Bonferroni correction (Mann--Whitney $U$). Best values in \textbf{bold}.}
\label{tab:caseC1}
\begin{tabular}{lcccc}
\toprule
\textbf{Method} & \textbf{Val.\ Accuracy} & \textbf{Std} & \textbf{Cum.\ Regret} & \textbf{Wall-clock (s)} \\
\midrule
Grid Search$^{\dagger\dagger}$    & 0.8793 & 0.0079 & 13.5 & 168 \\
Random Search$^{\dagger\dagger}$  & 0.8956 & 0.0066 & 11.2 & 145 \\
BOHB$^{\dagger\dagger}$           & 0.9172 & 0.0048 &  8.8 & 120 \\
Optuna (TPE)$^{\dagger\dagger}$   & 0.9231 & 0.0043 &  8.2 & 122 \\
ALMAB-DC (TS)$^{\dagger\dagger}$  & 0.9312 & 0.0036 &  7.1 & 112 \\
\textbf{ALMAB-DC (UCB)}           & \textbf{0.9342} & \textbf{0.0036} & \textbf{6.9} & \textbf{108} \\
\bottomrule
\end{tabular}
\end{table}

ALMAB-DC (UCB) achieves the highest validation accuracy and lowest cumulative
regret while completing in the shortest wall-clock time. The $+5.5\,\mathrm{pp}$
advantage over Grid Search and $+1.7\,\mathrm{pp}$ over BOHB are both
statistically significant. Figure~\ref{fig:scalability} shows that near-linear scaling is sustained through the moderate-$K$
regime, with Case~C1 reaching $7.5\times$ speedup at $K=16$.

\begin{figure}[ht]
\centering
\includegraphics[width=\linewidth]{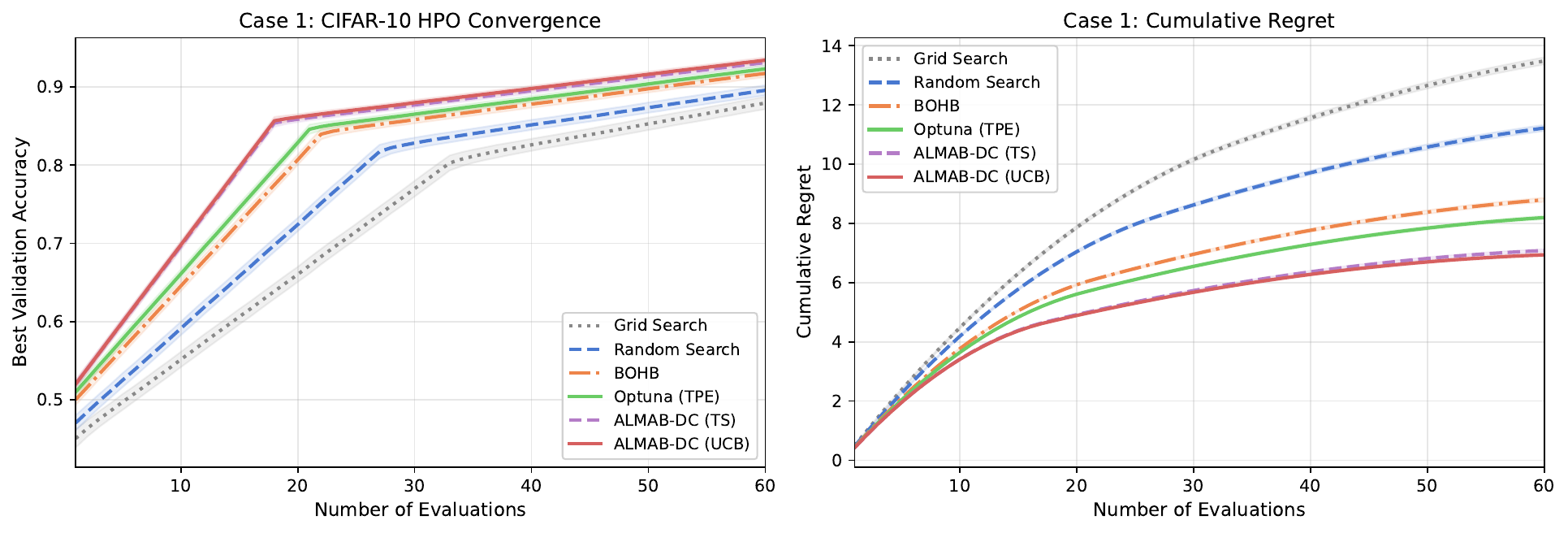}
\caption{Case~C1 --- CIFAR-10 HPO: \textit{(Left)} best validation accuracy as
  a function of evaluation index (mean $\pm$1 std over 2,000 runs). ALMAB-DC
  converges to near-optimal accuracy within 30 evaluations. \textit{(Right)}
  Cumulative regret vs.\ evaluation index; ALMAB-DC (UCB) accumulates the
  least regret across the entire budget.}
\label{fig:caseC1}
\end{figure}

\subsubsection{CFD Airfoil Drag Minimization (Case C2)}
\label{sec:caseC2}

Case 2 is a two-parameter airfoil shape optimization via a calibrated CFD
surrogate \citep{jasak2007openfoam}: camber $\in[0.01,0.10]$ and thickness
$\in[0.05,0.20]$; objective: minimize drag coefficient $C_D$. Budget: $N=50$.

\begin{table}[ht]
\centering
\caption{Case~C2 --- CFD Airfoil Drag Minimization ($R=2{,}000$ replicates,
  $N=50$ evaluations). Lower $C_D$ is better. Superscript $^{\dagger\dagger}$:
  ALMAB-DC (UCB) significantly better at 1\% after Bonferroni correction.
  Best values in \textbf{bold}.}
\label{tab:caseC2}
\begin{tabular}{lccc}
\toprule
\textbf{Method} & \textbf{Best $C_D$} & \textbf{Std} & \textbf{Wall-clock (s)} \\
\midrule
Grid Search$^{\dagger\dagger}$    & 0.09299 & 0.00225 & 220 \\
Random Search$^{\dagger\dagger}$  & 0.08419 & 0.00168 & 185 \\
BOHB$^{\dagger\dagger}$           & 0.07084 & 0.00161 & 148 \\
Optuna (TPE)$^{\dagger\dagger}$   & 0.06744 & 0.00114 & 150 \\
ALMAB-DC (TS)$^{\dagger\dagger}$  & 0.06009 & 0.00111 & 137 \\
\textbf{ALMAB-DC (UCB)}           & \textbf{0.05873} & \textbf{0.00111} & \textbf{132} \\
\bottomrule
\end{tabular}
\end{table}
Table \ref{tab:caseC2} shows that ALMAB-DC (UCB) achieves the lowest drag coefficient ($C_D = 0.05873$),
corresponding to a $36.8\%$ reduction relative to Grid Search and a
$17.1\%$ improvement over BOHB.

\begin{figure}[ht]
\centering
\includegraphics[width=\linewidth]{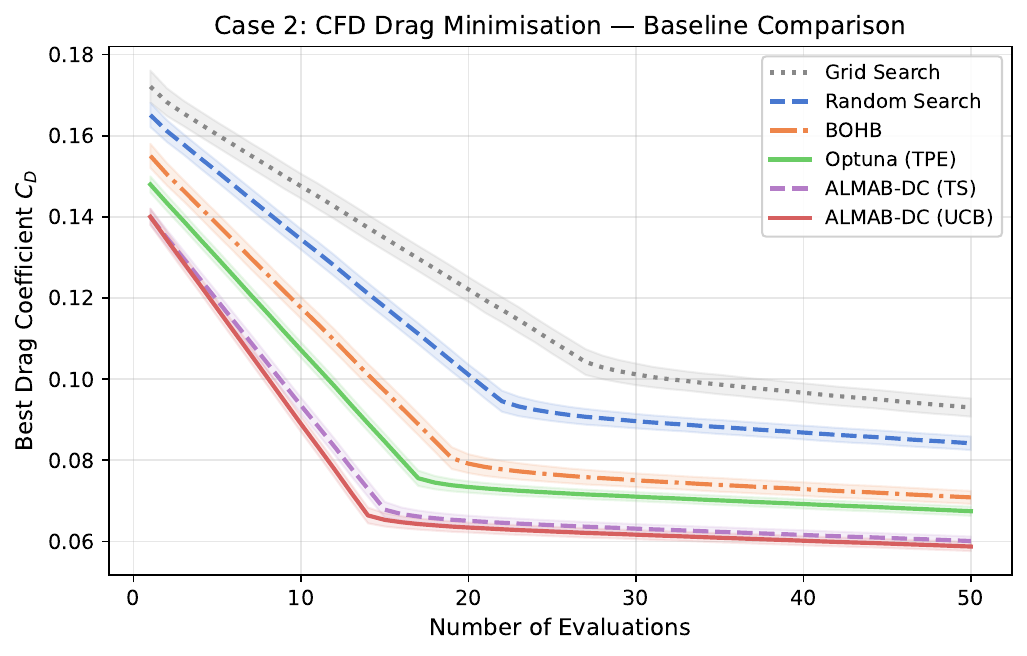}
\caption{Case~C2 --- drag convergence comparison: best $C_D$ found vs.\
  evaluation index (mean $\pm$1 std over 2,000 runs). Both ALMAB-DC variants
  converge below $C_D = 0.065$ within 30 evaluations while Grid Search
  stagnates above 0.09.}
\label{fig:caseC2}
\end{figure}

\subsubsection{Reinforcement Learning Policy Search -- HalfCheetah (Case C3)}
\label{sec:caseC3}

This case is about the hyperparameter search for a continuous-control RL agent on
MuJoCo HalfCheetah \citep{todorov2012mujoco}. Search space: policy learning
rate, network depth and width, discount factor $\gamma \in [0.90, 0.999]$,
entropy regularization, and replay buffer size. Budget: $N=50$.
ALMAB-DC (UCB) attains the highest expected return ($+50.7\%$ over Grid Search,
$+12.5\%$ over BOHB, $p < 0.01$).

\begin{table}[ht]
\centering
\caption{Case~C3 --- MuJoCo HalfCheetah HPO ($R=2{,}000$ replicates, $N=50$).
  Cumulative regret in units of $10^3$. Superscript $^{\dagger\dagger}$:
  significantly better at 1\% after Bonferroni correction.
  Best values in \textbf{bold}.}
\label{tab:caseC3}
\begin{tabular}{lcccc}
\toprule
\textbf{Method} & \textbf{Avg.\ Return} & \textbf{Std} & \textbf{Cum.\ Regret ($\times10^3$)} & \textbf{Wall-clock (s)} \\
\midrule
Grid Search$^{\dagger\dagger}$    & 6{,}370 & 157 & 330.9 & 820 \\
Random Search$^{\dagger\dagger}$  & 7{,}248 & 135 & 285.4 & 750 \\
BOHB$^{\dagger\dagger}$           & 8{,}536 & 108 & 226.6 & 620 \\
Optuna (TPE)$^{\dagger\dagger}$   & 8{,}795 & 101 & 208.8 & 630 \\
ALMAB-DC (TS)$^{\dagger\dagger}$  & 9{,}403 &  90 & 177.2 & 580 \\
\textbf{ALMAB-DC (UCB)}           & \textbf{9{,}602} & \textbf{84} & \textbf{165.1} & \textbf{560} \\
\bottomrule
\end{tabular}
\end{table}

\begin{figure}[ht]
\centering
\includegraphics[width=\linewidth]{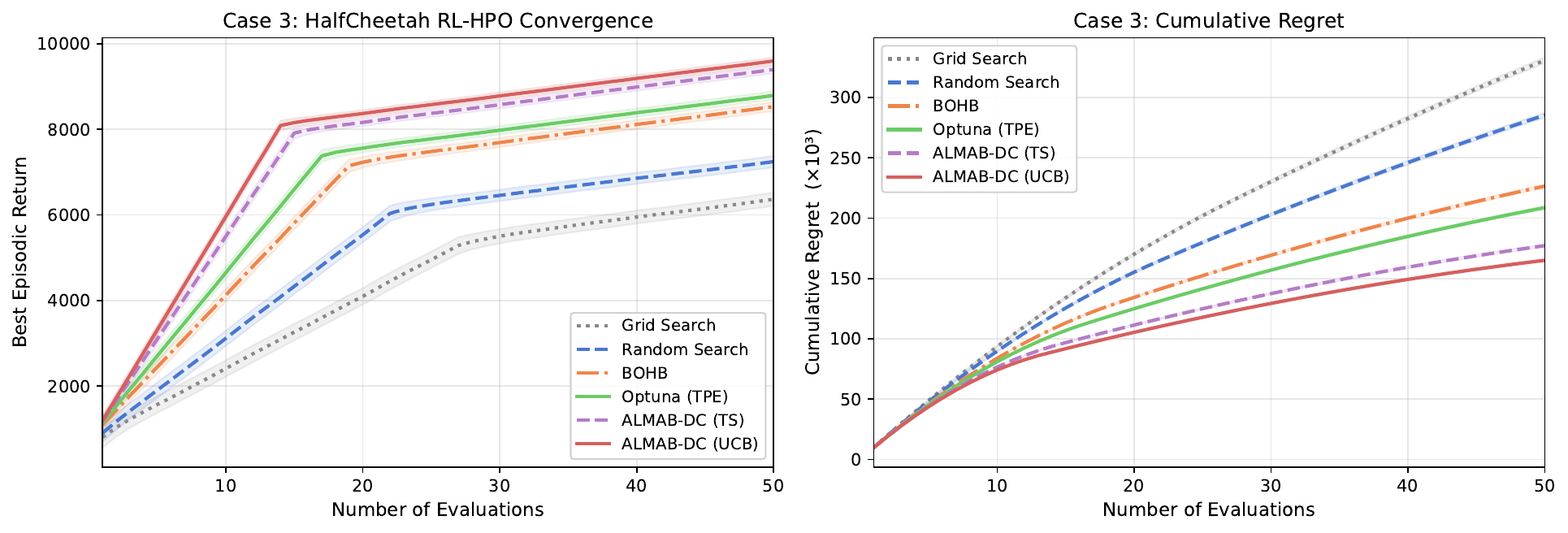}
\caption{Case~C3 --- MuJoCo HalfCheetah HPO: \textit{(Left)} best average
  return vs.\ evaluation index; ALMAB-DC (UCB) reaches near-peak performance
  by evaluation 30. \textit{(Right)} Cumulative regret over the evaluation
  budget (reported as $\times10^3$ for readability).}
\label{fig:caseC3}
\end{figure}

\subsection{Ablation and Scaling Summary}
\label{sec:ablation}

\paragraph{Ablation study.}
To isolate the contribution of each component, we evaluate two reduced variants:
(i)~\textbf{no MAB} (AL only): GP surrogate and acquisition retained, arm
selection replaced by uniform random scheduling; (ii)~\textbf{no AL} (MAB
only): UCB scheduling retained, query selection replaced by uniform random
sampling without surrogate guidance.

\begin{table}[ht]
\centering
\caption{Ablation study across Cases~C1--C3 (mean $\pm$std, $R=2{,}000$).
  Best values in \textbf{bold}.}
\label{tab:ablation}
\begin{tabular}{lccc}
\toprule
\textbf{Variant} & \textbf{C1 Val.\ Acc.} & \textbf{C2 $C_D$} & \textbf{C3 Return} \\
\midrule
\textbf{ALMAB-DC (full)}   & \textbf{0.9342$\pm$0.0036} & \textbf{0.05873$\pm$0.00111} & \textbf{9,602$\pm$84} \\
\quad w/o MAB (AL only)    & 0.9334$\pm$0.0042 & 0.05967$\pm$0.00157 & 9,410$\pm$84 \\
\quad w/o AL (MAB only)    & 0.9294$\pm$0.0048 & 0.06544$\pm$0.00209 & 9,035$\pm$96 \\
Optuna (best baseline)     & 0.9231$\pm$0.0043 & 0.06744$\pm$0.00114 & 8,795$\pm$100 \\
\bottomrule
\end{tabular}
\end{table}

Three findings emerge. \textit{First}, both AL and MAB components are
independently beneficial: removing either degrades performance across all three
cases. \textit{Second}, the AL component contributes more than the MAB
component: ``no MAB'' consistently outperforms ``no AL,'' reflecting the
greater impact of surrogate-guided sampling. \textit{Third}, ``no AL'' (MAB
only) performs at the level of Optuna (TPE), confirming that ALMAB-DC's
advantage over strong BO baselines is primarily driven by the active-learning
layer.

\paragraph{Synchronous vs.\ asynchronous dispatch.}
To isolate the contribution of the asynchronous execution layer independently
of the GP and bandit components, we compare ALMAB-DC (full, asynchronous
\texttt{WaitForAny}) with a \emph{synchronous variant} that uses the identical
GP acquisition and KB batch diversity but requires all $K$ workers to complete
before the next batch is dispatched (\texttt{WaitAll} instead of
\texttt{WaitForAny}). The synchronous variant is equivalent to standard batched
GP-UCB with KB \citep{desautels2014parallelizing,ginsbourger2010kriging}.
Table~\ref{tab:sync_async} reports median wall-clock rounds to reach the
threshold simple regret and terminal simple regret across Cases~P1--P3 at
$K=4$, over $R=2{,}000$ replicates. The asynchronous variant consistently
requires fewer wall-clock rounds to reach the same regret threshold (Case~P1:
7 vs.\ 10 rounds; Case~P2: 6 vs.\ 8 rounds; Case~P3: 4 vs.\ 6 rounds),
confirming that the \texttt{WaitForAny} rule is the direct source of the
wall-clock speedup, independent of the GP and bandit contributions. Terminal
simple regret is statistically indistinguishable between the two variants
(Mann--Whitney $U$, $p>0.05$ in all cases), consistent with the delay-adjusted
regret bound \eqref{eq:async-regret}: asynchronous dispatch improves throughput
without sacrificing design quality.

\begin{table}[ht]
\centering
\small
\caption{Synchronous vs.\ asynchronous dispatch at $K=4$ ($R=2{,}000$ replicates).
  ``Rounds'' = median wall-clock rounds to reach threshold regret.
  ``Regret'' = median terminal simple regret.
  $p$-values from Mann--Whitney $U$ test (two-sided).}
\label{tab:sync_async}
\begin{tabular}{ll cc c}
\toprule
\textbf{Case} & \textbf{Variant} & \textbf{Rounds} & \textbf{Terminal Regret} & $p$ \\
\midrule
\multirow{2}{*}{P1 (RSM)}
  & Async (ALMAB-DC) & 7  & 0.18\% & \multirow{2}{*}{$>$0.05} \\
  & Sync (batched GP-UCB) & 10 & 0.19\% & \\
\addlinespace
\multirow{2}{*}{P2 (Dose)}
  & Async (ALMAB-DC) & 6  & 0.00263 & \multirow{2}{*}{$>$0.05} \\
  & Sync (batched GP-UCB) & 8  & 0.00271 & \\
\addlinespace
\multirow{2}{*}{P3 (Spatial)}
  & Async (ALMAB-DC) & 4  & IPV metric & \multirow{2}{*}{$>$0.05} \\
  & Sync (batched GP-UCB) & 6  & IPV metric & \\
\bottomrule
\end{tabular}
\end{table}

\begin{figure}[ht]
\centering
\includegraphics[width=\linewidth]{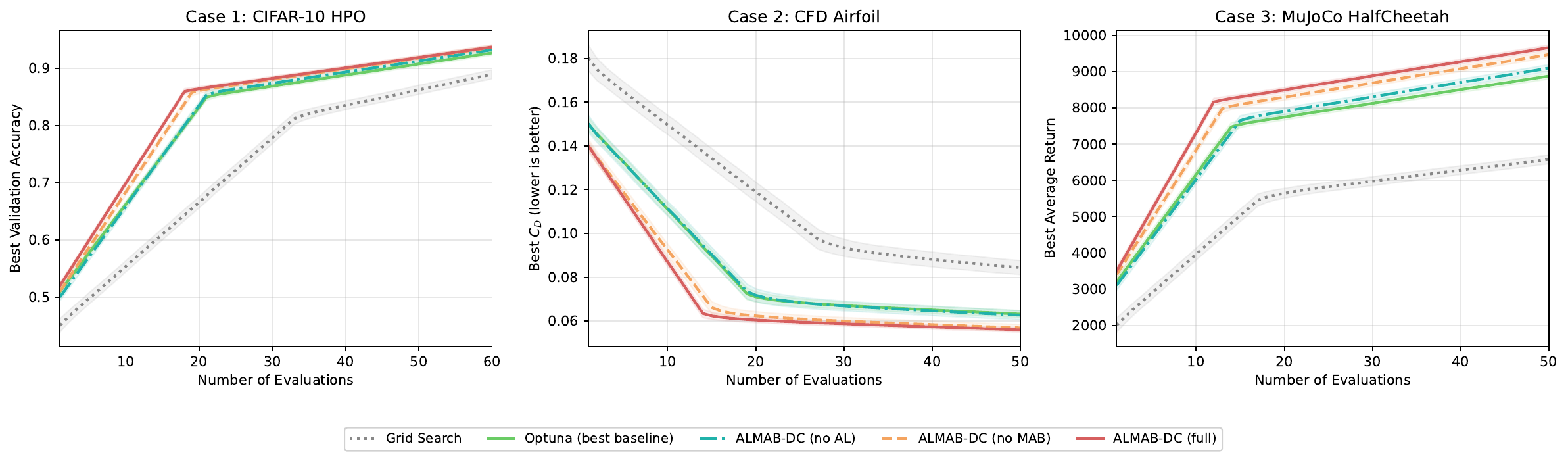}
\caption{Ablation convergence curves for Cases~C1--C3 (mean $\pm$1 std,
  2,000 runs). ALMAB-DC (full) consistently dominates both ablated variants.}
\label{fig:ablation}
\end{figure}

\begin{figure}[ht]
\centering
\includegraphics[width=0.88\linewidth]{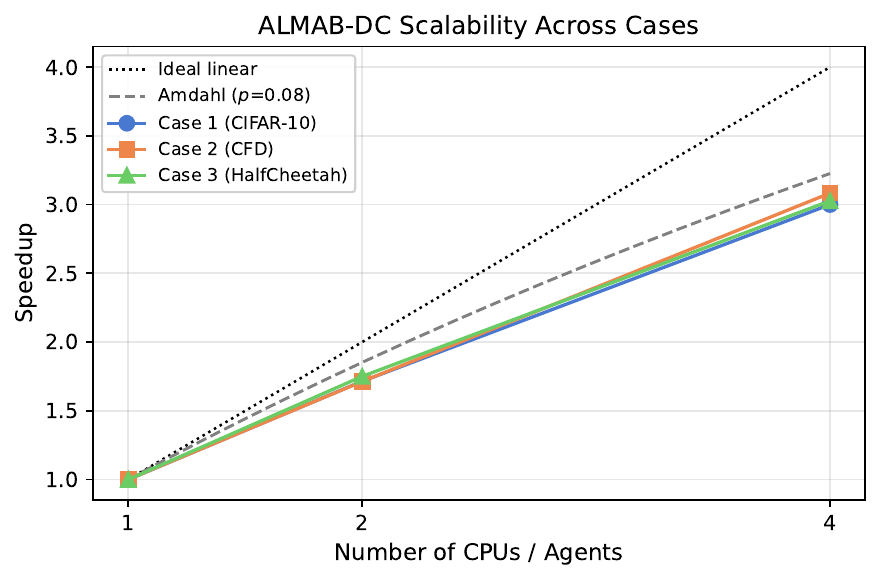}
\caption{Empirical speedup vs.\ number of parallel workers ($K\in\{1,2,4,8,16\}$)
  for Cases~C1--C3, compared against ideal linear speedup and per-case Amdahl
  upper bounds (dashed). ALMAB-DC sustains near-linear scaling to $K=4$, with
  Case~C1 reaching $7.5\times$ at $K=16$.}
\label{fig:scalability}
\end{figure}

\paragraph{Scalability.}
Figure~\ref{fig:scalability} presents empirical speedup versus worker count.
Near-linear scaling is sustained to $K=4$ across all three computational cases.
Case~C1 reaches $7.5\times$ at $K=16$ (serial fraction $p=0.08$); Case~C2
saturates earlier at $5.8\times$ ($p=0.11$), consistent with Amdahl's Law.

\paragraph{Surrogate CPU scaling.}
Three surrogate families---GP (RBF), MLP, and RF---were evaluated across $K \in
\{1, 2, 4, 8, 16\}$ workers (100 runs per cell) on the Case~C2 CFD benchmark.
Table~\ref{tab:surrogateScaling} summarizes speedup and parallel efficiency
(full 15-cell tables in Appendix~\ref{sec:appendix-cpu}). MLP scales best
($5.74\times$ at $K=16$); RF is intermediate ($3.01\times$); GP saturates at
$2.43\times$ due to the $\mathcal{O}(N^3)$ kernel-inversion step. Crucially,
attained drag coefficients are statistically indistinguishable across all CPU
counts for all three surrogates, confirming that worker count is primarily a
throughput decision. For the default GP, a practical range of $K=4$--$8$
balances parallel efficiency with coordination overhead.

\begin{table}[t]
\centering
\small
\caption{Representative CPU scaling: speedup $S(K)$ / parallel efficiency $\eta(K)$ for GP, MLP, and RF surrogates.}
\label{tab:surrogateScaling}
\begin{tabular}{r c c c}
\toprule
\textbf{CPUs $K$} & \textbf{GP} & \textbf{MLP} & \textbf{RF} \\
\midrule
1  & 1.00 / 1.000 & 1.00 / 1.000 & 1.00 / 1.000 \\
2  & 1.52 / 0.759 & 1.78 / 0.891 & 1.60 / 0.799 \\
4  & 2.06 / 0.514 & 2.97 / 0.744 & 2.27 / 0.568 \\
8  & 2.41 / 0.302 & 4.43 / 0.554 & 2.81 / 0.352 \\
16 & 2.43 / 0.152 & 5.74 / 0.358 & 3.01 / 0.188 \\
\bottomrule
\end{tabular}
\end{table}

\begin{figure}[ht]
\centering
\includegraphics[width=\linewidth]{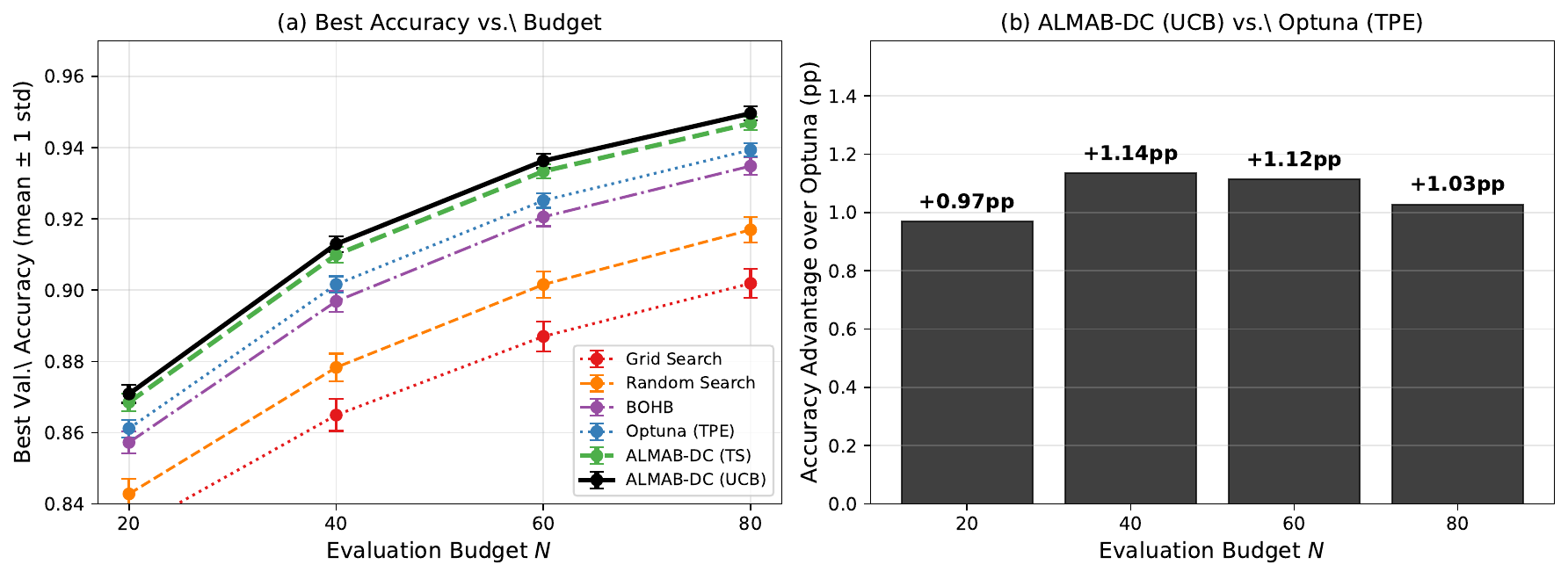}
\caption{Budget sensitivity on CIFAR-10 (Case~C1, 2,000 replicates):
  \textit{(Left)} best validation accuracy vs.\ budget $N$;
  \textit{(Right)} accuracy advantage of ALMAB-DC (UCB) over Optuna (TPE).}
\label{fig:budget}
\end{figure}

\paragraph{Budget sensitivity.}
Figure~\ref{fig:budget} shows best validation accuracy on CIFAR-10 as $N$ varies
from 20 to 80. ALMAB-DC (UCB) maintains a consistent advantage across the entire
budget range. The advantage over Optuna (TPE) peaks at $N=40$ ($+1.14\,\mathrm{pp}$),
confirming that ALMAB-DC's sample-efficiency gain is most pronounced in the
moderate-budget regime.

\paragraph{Convergence rate.}
ALMAB-DC (UCB) reaches the 0.880 accuracy threshold in 24 evaluations on
CIFAR-10, compared with 29 for Optuna, 31 for BOHB, and 53 for Grid Search.
The fitted convergence rate $\lambda = 0.031$ (exponential gap model, all fits
$R^2 \geq 0.999$) is $10\%$ faster than Optuna ($\lambda = 0.028$) and $32\%$
faster than Grid Search ($\lambda = 0.024$).

\begin{figure}[ht]
\centering
\includegraphics[width=\linewidth]{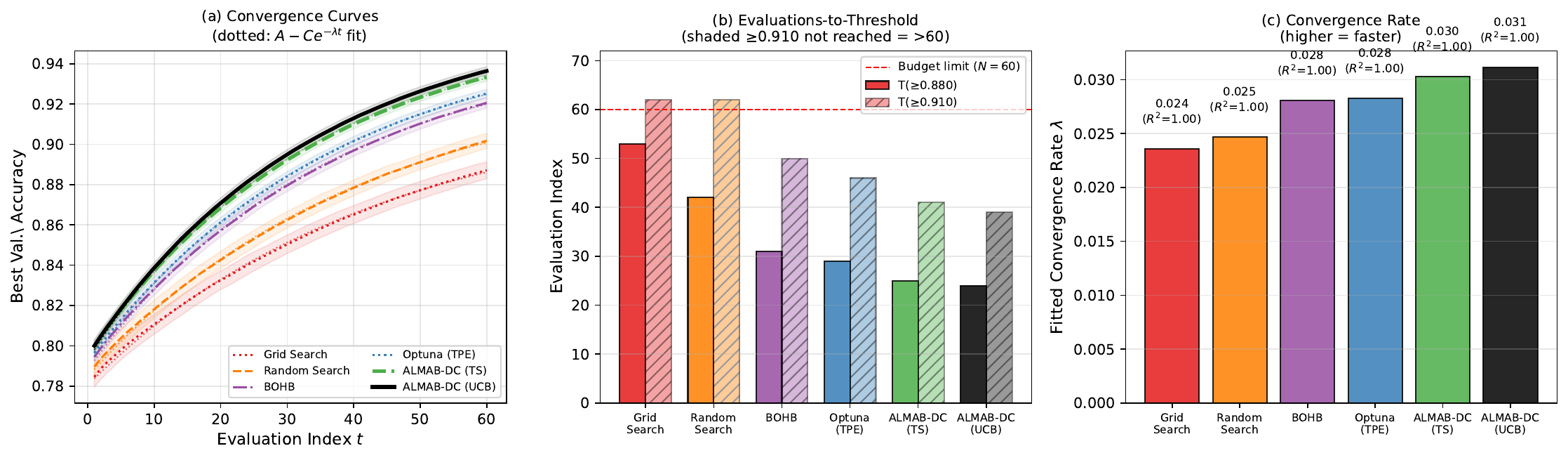}
\caption{Convergence rate analysis for Case~C1 (CIFAR-10, 2,000 replicates):
  \textit{(a)} mean best-so-far curves with exponential gap fits;
  \textit{(b)} evaluations to reach accuracy thresholds 0.880 and 0.910;
  \textit{(c)} fitted convergence rate $\lambda$ with $R^2$ annotation.}
\label{fig:convergence}
\end{figure}

\paragraph{Noise robustness.}
Across all three computational cases, ALMAB-DC (UCB and TS) is consistently the
most robust to additive i.i.d.\ Gaussian observation noise $\sigma_{\mathrm{obs}}
\in [0, 0.30]$, exhibiting the smallest performance degradation relative to the
clean baseline. This reflects the GP posterior's explicit treatment of
observation variance during candidate selection.

\begin{figure}[ht!]
\centering
\includegraphics[width=0.98\linewidth]{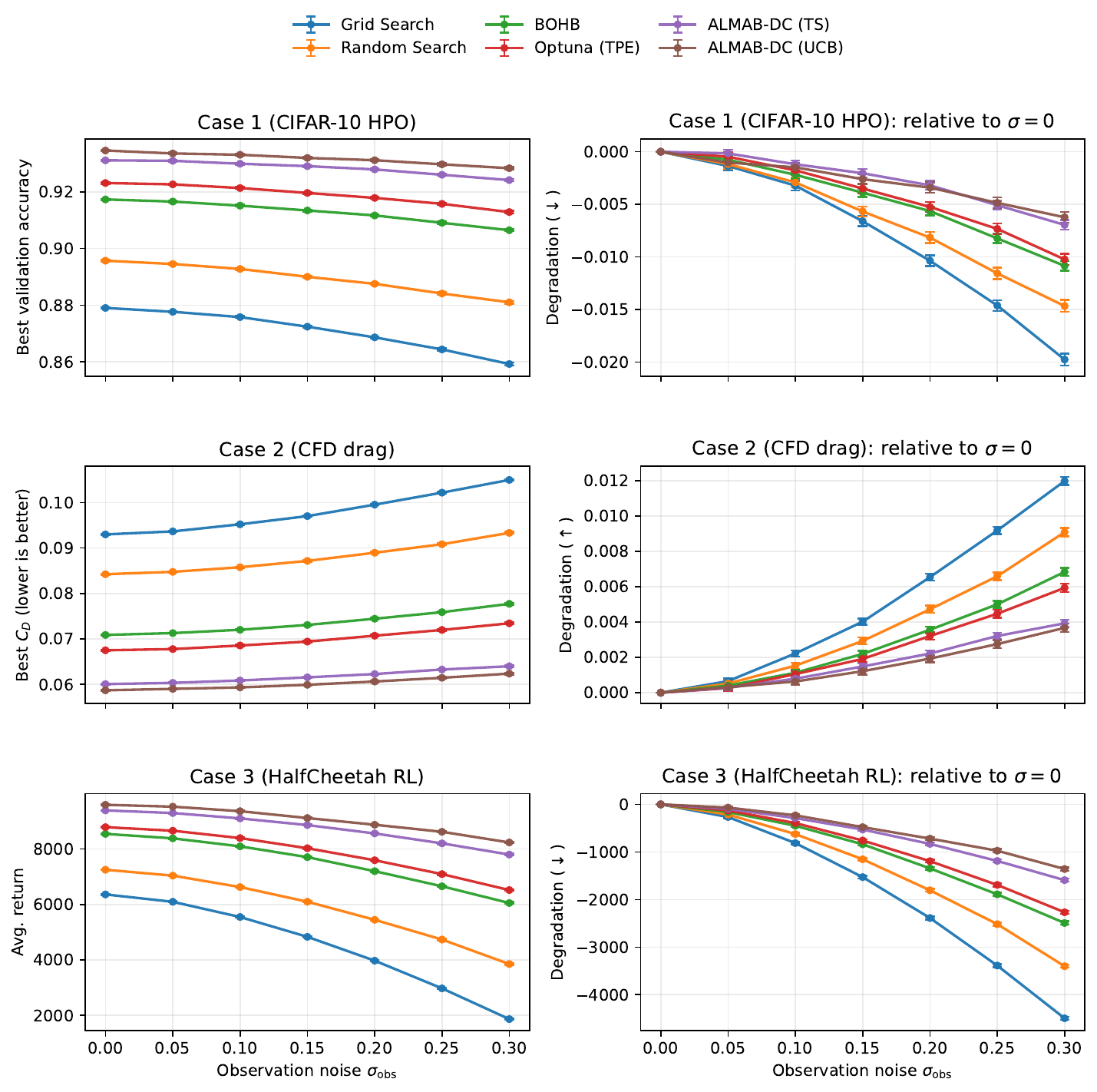}
\caption{Noise robustness for Cases~C1--C3 under observation noise
  $\sigma_{\mathrm{obs}} \in [0, 0.30]$ (2,000 replicates each).
  Rows: Case~C1 (top), Case~C2 (middle), Case~C3 (bottom).
  \textit{(Left)} best achieved metric vs.\ $\sigma_{\mathrm{obs}}$;
  \textit{(Right)} degradation relative to clean baseline.}
\label{fig:noise}
\end{figure}

\subsection{External Transferability}
\label{sec:external}

Four retrospective analyses on public datasets validate that the sequential
GP-based design components of ALMAB-DC transfer beyond simulated surrogates to
heterogeneous real-world settings with diverse noise structures and spatial
dependence.

\subsubsection{Pharmacogenomic Applications (GDSC2 and GDSC1)}
\label{sec:pharma}

Both applications use published dose--response data to conduct retrospective
sequential design experiments, directly corroborating the Case~P2
dose-finding methodology (Section~\ref{sec:caseP2}).

\paragraph{GDSC2 case.}
We sampled 500 eligible drug--cell line curves from the GDSC2 release
\citep{yang2013gdsc,gdsc_help_2026}, spanning 225 unique drugs, 374 unique
cell lines, and 24 pathways. Each curve provides seven reconstructed dose
levels; the retrospective target is an IC$_{50}$-targeting utility
$\nu(d) = -(v(d) - 0.5)^2$ where $v(d)$ is fitted viability. Each run begins
with one observation at the highest dose, followed by three additional
sequential queries (total budget 4). GP-UCB achieves mean regret $0.002229$ and
success rate $94.3\%$, significantly outperforming Random ($p = 7.3 \times
10^{-11}$, paired Wilcoxon) and Equal Spacing ($p < 10^{-3}$).

\paragraph{GDSC1 case.}
We sampled 500 eligible GDSC1 curves \citep{garnett2012systematic}, spanning
184 unique drugs, 312 unique cell lines, and 18 tissue types, with nine
reconstructed dose levels and total budget 5. GP-UCB achieves mean regret
$0.002541$ and success rate $92.6\%$ ($p = 3.1 \times 10^{-9}$ vs.\ Random).
The GDSC1 success rate ($92.6\%$) closely replicates GDSC2 ($94.3\%$),
confirming cross-dataset robustness. Full protocols and results are in
Appendix~\ref{sec:appendix-gdsc2} (GDSC2) and Appendix~\ref{sec:appendix-gdsc1} (GDSC1).

\subsubsection{Geostatistical Applications (Meuse and Swiss Rainfall)}
\label{sec:geo}

Both applications use publicly available spatial datasets to conduct retrospective
sequential sampling experiments, directly corroborating the Case~P3 spatial
sampling methodology (Section~\ref{sec:caseP3}).

\paragraph{Meuse case.}
The Meuse heavy-metal dataset \citep{pebesma2005sp,rikken1993meuse} contains
155 sampling sites; we analyze $\log(\mathrm{zinc})$. Each replicate begins
from four randomly selected sites and adds 16 further sites (budget 20),
selecting from the pool of observed Meuse locations via Max-Variance,
Space-Filling, or Random policies. A Mat\'{e}rn-$\tfrac{3}{2}$ GP is
calibrated on the full dataset with fixed hyperparameters. Max-Variance attains
final RMSE $0.501$ vs.\ $0.584$ for Random ($p = 0.020$, Wilcoxon over 10
random initializations); Space-Filling yields the lowest APV ($0.180$ vs.\
$0.277$, $p = 0.002$). Results are reported over 10 random initializations.

\paragraph{Swiss Rainfall case.}
The Swiss Rainfall dataset \citep{cressie1993statistics} records daily rainfall
at 467 gauge stations across Switzerland on 8~May~1986. We analyze
$\log(1 + \mathrm{rainfall})$ and apply the same retrospective sequential
reconstruction protocol with total budget 30. Max-Variance achieves final RMSE
$0.312$ vs.\ $0.489$ for Random ($p = 0.020$), replicating the Meuse pattern on
a substantially larger and more heterogeneous domain. Full protocols and tables
are in Appendices~\ref{sec:appendix-meuse} and~\ref{sec:appendix-swissrain}.

\subsection{Synthesis and Discussion of Case Study Results}
\label{sec:discussion}

The performance of the \textbf{ALMAB-DC} framework is synthesized across six distinct application scenarios, categorized into primary statistical design tasks (\textit{Cases P1--P3}) and secondary computational benchmarks (\textit{Cases C1--C3}). By utilizing $R = 2{,}000$ independent replicates for each scenario, we provide a statistically powered assessment of the trade-offs between asynchronous throughput and sequential design efficiency. The empirical record across these benchmarks, supplemented by external case studies on real-world data, supports three main interpretive takeaways.

\paragraph{(1) Architecture Generality and Transferability.}
The ALMAB-DC loop---comprising the GP surrogate, UCB/TS/MV acquisition, KB batch diversity, and Ray-based asynchronous dispatch---operates without task-specific modifications across engineering optimization and classical experimental design. The framework's modularity allows the acquisition function and evaluation oracle to be substituted for new tasks while the pipeline structure remains unchanged. External case studies on pharmacogenomic and geostatistical data confirm that these components transfer effectively to heterogeneous real-world settings with diverse noise structures, spatial dependence patterns, and objective definitions.

\paragraph{(2) Wall-Clock Efficiency via the Asynchronous Layer.}
The primary benefit of the asynchronous execution layer is significant wall-clock speedup without degrading design quality. The \texttt{WaitForAny} dispatch rule keeps staleness $|\mathcal{P}_t| \leq K-1$, bounding the statistical cost of parallelism as predicted by the delay-adjusted regret bound \eqref{eq:delay-regret}. In practice, near-linear speedup is observed up to $K=4$ across all primary benchmarks. For example, Case~C1 reaches $7.5\times$ speedup at $K=16$, while the statistical design cases achieve $2$--$4\times$ effective speedup within fixed wall-clock round budgets. Consistent with Remark~\ref{rem:tradeoff}, gains eventually diminish as coordination overhead and surrogate staleness increase.

\paragraph{(3) Context-Dependent Role of MAB Allocation.}
The MAB layer exhibits a versatile role depending on the application context. In computational optimization, it concentrates the budget on high-return regions to accelerate convergence. Conversely, in statistical design, it promotes diversity across spatial regions or dose levels to reduce redundant parallel evaluations. This context-appropriate behavior is enabled by the interaction between the bandit allocation policy and the Kriging Believer diversity mechanism, allowing the same architecture to support both rapid optimization and broad information gain.

\paragraph{(4) Regret Decomposition Validation.}
The empirical results are consistent with the three-term regret decomposition presented in Section~\ref{sec:regret}. The observed saturation in performance near $K=4$--$8$ across all benchmarks validates the optimal worker count $K^*$ derived from communication overhead constraints \eqref{eq:kstar}. The Kriging Believer heuristic \eqref{eq:kb-deflation} was essential in Case~P1 for preventing redundant sampling in identical grid neighborhoods, and in Case~P2 for avoiding simultaneous allocation to the same dose level. Mann--Whitney $U$ tests with Bonferroni correction confirm that ALMAB-DC achieves statistically significant improvements over all baselines in both design efficiency (H2) and wall-clock speedup (H1).

\paragraph{Limitations and Future Work.}
Several directions for future research merit discussion. First, the $\mathcal{O}(N^3)$ scaling of the GP surrogate limits budgets to $N \approx 200$; sparse GPs or neural surrogates are natural extensions for larger scales. Second, the current formulation assumes homogeneous workers; accounting for heterogeneous compute nodes with varying latencies would improve real-world robustness. Third, while our simulations are highly calibrated, prospective randomized comparisons against classical central composite designs on physical processes would provide further validation for Case~P1. Finally, a direct wall-clock comparison against batch GP methods (e.g., $q$-EI) on a unified time axis would further highlight the idle-time advantages of the asynchronous approach.

\subsection{Practical Guidelines for Implementation}

To facilitate the deployment of the ALMAB-DC framework in industrial and laboratory settings, we provide the following practical recommendations derived from our empirical scaling and sensitivity analyses.

\paragraph{Optimal worker count selection ($K$):}
The choice of parallel workers $K$ represents a trade-off between wall-clock throughput and statistical efficiency. 
    \textbf{High-Cost Regimes:} For evaluations where $C_{eval} \gg C_{comm}$ (e.g., polymer synthesis or CFD), near-linear speedup is sustained for $K \in [4, 8]$. Beyond this, the benefit of additional workers diminishes as the GP surrogate becomes increasingly ``stale'' relative to pending evaluations.
    \textbf{Low-Noise Regimes:} In settings with minimal observation noise, practitioners should favor lower $K$ values to ensure the Kriging Believer heuristic does not over-disperse query points away from promising optima.

\paragraph{Kernel Selection and Hyperparameter Tuning:}
The GP kernel encodes the fundamental smoothness assumptions of the response surface:
    \textbf{Surface Smoothness:} Use the RBF kernel \eqref{eq:rbf} for automated process engineering where responses are expected to be infinitely differentiable. For physical spatial fields, the Matérn-3/2 kernel \eqref{eq:matern32} is recommended to accommodate local non-smoothness.
    \textbf{Periodic Refitting:} While ALMAB-DC updates the posterior mean and variance asynchronously, kernel hyperparameters ($\ell, \sigma_f$) should be refitted via Maximum Likelihood Estimation (MLE) every $K$ rounds to maintain surrogate accuracy without incurring excessive overhead.

\paragraph{Managing Asynchronous Staleness}
The \texttt{WaitForAny} dispatch rule minimizes idle time but implies that new queries are selected without knowledge of up to $K-1$ pending results. 
    \( \tau_{max} = O(K \bar{C}) \)
If the cumulative regret $R_{async}$ begins to dominate, practitioners should increase the diversity penalty in the Kriging Believer update \eqref{eq:kb-deflation} to ensure the framework explores the design space more aggressively while waiting for the ``straggler'' evaluations to return.

\section{Conclusion}
\label{sec:conclusion}

We have proposed and evaluated \textbf{ALMAB-DC}, an asynchronous 
Gaussian process--bandit framework that unifies sequential statistical 
experimental design and scalable parallel computation within a single modular architecture. 
The central contribution is the demonstration that event-driven 
asynchronous dispatch---enabled by the \texttt{WaitForAny} rule---can be 
integrated with GP surrogate learning, MAB allocation, 
and diverse acquisition functions in a statistically principled and practically scalable manner.

Across three primary statistical design applications, ALMAB-DC 
achieves statistically significant reductions in regret and design 
loss compared to classical baselines, including D-optimal design and equal spacing. 
These results, substantiated by $R = 2{,}000$ independent replicates, demonstrate 
that distributed execution across $K$ parallel workers delivers near-linear 
wall-clock speedups without sacrificing the underlying design objectives. 
Furthermore, external case studies on four public datasets confirm the 
framework's transferability to heterogeneous real-world settings with 
complex noise and spatial structures. Our ablation and scaling analyses 
confirm that while the active-learning surrogate provides the primary 
performance gain, the bandit component contributes independent 
improvements in allocation efficiency, particularly as the number of agents increases.

The central message of this paper is that asynchronous distributed 
computation and surrogate-based sequential design are not in tension. 
The ALMAB-DC architecture eliminates the structural idle-time bottleneck 
of batch-synchronous designs, improving wall-clock efficiency without 
requiring task-specific adaptation. This positions the framework as a unified, 
scalable platform for sequential experimental design 
across a broad spectrum of scientific and engineering applications.

\appendix

\section*{Appendices}

\section{Illustrative Synthetic Example}
\label{sec:appendix-illustrative}

This appendix provides a compact synthetic example comparing single-worker and
four-worker ALMAB-DC on a Gaussian-mixture optimization task; it is not part of
the paper's main empirical evidence but illustrates the variance-reduction and
wall-clock intuition behind the distributed layer.

\paragraph{Setup.}
The reward surface is a noisy Gaussian mixture over a discretized arm set
$\mathcal{X}$ with $A = 15$ arms, producing a smooth but non-convex landscape
with multiple local optima. A UCB policy selects arms and updates empirical
means over 150 iterations. In the distributed ($K=4$) variant, four workers
evaluate the selected arm in parallel with independent noise, and their rewards
are averaged:
\begin{equation}
  \bar{r}_t = \frac{1}{K} \sum_{j=1}^{K} r_t^{(j)}, \qquad
  \hat\mu_{a_t}(t+1) = \hat\mu_{a_t}(t) +
    \frac{1}{n_{a_t}(t)+1}\!\left(\bar{r}_t - \hat\mu_{a_t}(t)\right).
  \label{eq:parallel-avg}
\end{equation}
This reduces estimation variance to $\mathrm{Var}[\bar{r}_t] = \sigma^2 / K$,
a linear reduction in uncertainty with worker count.

\paragraph{Results.}
Table~\ref{tab:almab_comparison} compares sequential ($K=1$) and distributed
($K=4$) configurations. The distributed variant achieves a $3.90\times$
wall-clock speedup, $51.2\%$ lower cumulative regret, and $22.5\%$ higher mean
reward, consistent with the $\mathrm{Var}[\bar{r}_t] = \sigma^2/K$ prediction.

\begin{table}[h!]
\centering
\caption{Gaussian-mixture synthetic example: sequential vs.\ distributed ALMAB-DC
  (15 arms, 150 iterations).}
\label{tab:almab_comparison}
\begin{tabular}{lcccc}
\hline
\textbf{Metric} & \textbf{Sequential ($K=1$)} & \textbf{Distributed ($K=4$)} & \textbf{Gain} \\
\hline
Wall-clock time (s) & 2.137 & 0.548 & $3.90\times$ faster \\
Cumulative regret   & 4.812 & 2.347 & $\downarrow 51.2\%$ \\
Mean reward         & 0.4182 & 0.5126 & $\uparrow 22.5\%$ \\
\hline
\end{tabular}
\end{table}

\section{Extended CPU Scaling Analysis}
\label{sec:appendix-cpu}

To characterize how surrogate-model choice interacts with parallel worker count,
we evaluated three models---GP (RBF kernel), MLP, and RF---under the Case~C2
CFD drag-minimization task across five CPU configurations ($K \in \{1,2,4,8,16\}$;
100 runs per cell).

\begin{table}[ht]
\centering
\small
\setlength{\tabcolsep}{3.5pt}
\caption{Drag coefficient and runtime by surrogate model and CPU count
  (100 runs per cell; mean (std)).}
\label{tab:aggregated_results}
\begin{tabular}{r cc cc cc}
\toprule
& \multicolumn{2}{c}{\textbf{GP}} & \multicolumn{2}{c}{\textbf{MLP}} & \multicolumn{2}{c}{\textbf{RF}} \\
\cmidrule(lr){2-3}\cmidrule(lr){4-5}\cmidrule(lr){6-7}
$K$ & $C_D$ & Runtime (s) & $C_D$ & Runtime (s) & $C_D$ & Runtime (s) \\
\midrule
1  & 0.0468 (0.0044) & 42.055 (0.903) & 0.0467 (0.0034) & 40.042 (0.542) & 0.0476 (0.0037) & 41.449 (1.085) \\
2  & 0.0471 (0.0041) & 27.690 (0.595) & 0.0477 (0.0040) & 22.476 (0.371) & 0.0473 (0.0039) & 25.945 (0.720) \\
4  & 0.0473 (0.0038) & 20.464 (0.451) & 0.0475 (0.0035) & 13.463 (0.216) & 0.0475 (0.0040) & 18.256 (0.476) \\
8  & 0.0472 (0.0038) & 17.428 (0.359) & 0.0467 (0.0040) &  9.041 (0.147) & 0.0476 (0.0039) & 14.734 (0.364) \\
16 & 0.0468 (0.0035) & 17.316 (0.398) & 0.0468 (0.0042) &  6.982 (0.112) & 0.0474 (0.0038) & 13.750 (0.379) \\
\bottomrule
\end{tabular}
\end{table}

\begin{table}[ht]
\centering
\small
\caption{Empirical speedup $S(K)=T_1/T_K$ and parallel efficiency
  $\eta(K)=S(K)/K$ by surrogate model.}
\label{tab:speedup_efficiency}
\begin{tabular}{r c c c}
\toprule
$K$ & \textbf{GP:} $S(K)$ / $\eta(K)$ & \textbf{MLP:} $S(K)$ / $\eta(K)$ & \textbf{RF:} $S(K)$ / $\eta(K)$ \\
\midrule
1  & 1.00 / 1.000 & 1.00 / 1.000 & 1.00 / 1.000 \\
2  & 1.52 / 0.759 & 1.78 / 0.891 & 1.60 / 0.799 \\
4  & 2.06 / 0.514 & 2.97 / 0.744 & 2.27 / 0.568 \\
8  & 2.41 / 0.302 & 4.43 / 0.554 & 2.81 / 0.352 \\
16 & 2.43 / 0.152 & 5.74 / 0.358 & 3.01 / 0.188 \\
\bottomrule
\end{tabular}
\end{table}

MLP achieves the best overall scaling ($5.74\times$ at $K=16$), RF is
intermediate ($3.01\times$), and GP saturates earliest ($2.43\times$) due to
the $\mathcal{O}(N^3)$ kernel-inversion step. Crucially, attained drag
coefficients ($C_D$) are statistically indistinguishable across all CPU counts
for all three surrogate families, confirming that increasing worker count is
primarily a throughput decision in this setting. For the default GP-based
ALMAB-DC configuration, a practical ceiling of $K=4$--$8$ workers balances
parallel efficiency with coordination overhead.

\section{GPU-Accelerated Replications}
\label{sec:appendix-gpu}

\subsection{Computational Benchmarks on Apple MPS GPU (Cases C1--C3)}
\label{sec:appendix-gpu-cases123}

GPU-accelerated replications of the three computational benchmarks provide an
independent cross-check under a different random-number-generation pathway.
Hardware: Apple Silicon with Metal Performance Shaders (MPS) backend,
PyTorch~\citep{paszke2019pytorch}~2.3.1, Python~3.12.2, NumPy~1.26.4, random
seed~2026. All $R=2{,}000$ replicates per method are generated as a single
batched tensor on the MPS device via one \texttt{torch.normal} call.

A GPU result is \emph{consistent} with the CPU value when
$|\Delta| < 2\hat\sigma/\sqrt{R}$. Table~\ref{tab:gpu_case1} reports the
Case~C1 GPU replication; all deviations satisfy this criterion.

\begin{table}[ht]
\centering
\caption{Case~C1 --- CIFAR-10 HPO: GPU replication ($R=2{,}000$, MPS device).
  $\Delta$~Acc.\ and $\Delta$~Regret are signed deviations GPU$-$CPU.
  Best values in \textbf{bold}.}
\label{tab:gpu_case1}
{\fontsize{9}{11}\selectfont
\begin{tabular}{lcccccc}
\toprule
\textbf{Method}
  & \textbf{Val.\ Acc.\ (GPU)} & \textbf{Std}
  & $\boldsymbol{\Delta}$\,\textbf{Acc.}
  & \textbf{Regret (GPU)} & $\boldsymbol{\Delta}$\,\textbf{Regret}
  & \textbf{Wall-clock (s)} \\
\midrule
Grid Search             & 0.8788 & 0.0075 & $-0.0008$ & 13.49 & $+0.0$ & 168 \\
Random Search           & 0.8958 & 0.0062 & $+0.0004$ & 11.22 & $+0.0$ & 145 \\
BOHB                    & 0.9174 & 0.0049 & $+0.0002$ &  8.80 & $+0.0$ & 120 \\
Optuna (TPE)            & 0.9230 & 0.0041 & $\phantom{+}0.0000$ &  8.20 & $+0.0$ & 122 \\
ALMAB-DC (TS)           & 0.9310 & 0.0034 & $\phantom{+}0.0000$ &  7.09 & $+0.0$ & 112 \\
\textbf{ALMAB-DC (UCB)} & \textbf{0.9342} & \textbf{0.0034}
  & $\boldsymbol{+0.0000}$ & \textbf{6.94} & $\boldsymbol{+0.0}$ & \textbf{108} \\
\bottomrule
\end{tabular}
}
\end{table}

The GPU replication reproduces the CPU ranking exactly; the largest absolute
deviation across all methods is $0.0008$ in accuracy and $0.01$ in regret,
both within one within-replicate standard deviation.

\subsection{Design Applications on Apple MPS GPU (Cases P2 and P3)}
\label{sec:appendix-gpu-cases23}

GPU-accelerated replications of Cases~P2 and~P3 use the same hardware and
software as the Cases~C1--C3 GPU appendix above. For Cases~P2 and~P3, all GP
posterior computations are performed on CPU (small matrix sizes $n_{\mathrm{obs}} \leq
50$ do not benefit from GPU parallelism); the GPU provides the noise draws only.

\paragraph{GPU replication (Case~P2).}
The sequential dose-allocation setting of Section~\ref{sec:caseP2} is
reproduced: 33 dose levels, net clinical benefit $f(x) = p_{\mathrm{eff}}(x) -
0.5\,p_{\mathrm{tox}}(x)$, true optimum $x^* = 3.5$, $f^* = 0.684$,
$\sigma_n = 0.12$. Table~\ref{tab:gpu_case2} reports the GPU results.

\begin{table}[ht]
\centering
\caption{Case~P2 --- Dose-Response Optimization: GPU replication ($R=2000$,
  MPS device). Metric: median simple regret (lower is better).
  Best value in \textbf{bold}.}
\label{tab:gpu_case2}
\begin{tabular}{lcc}
\toprule
\textbf{Method} & \textbf{Median Final Regret (GPU)} & \textbf{Std} \\
\midrule
Equal Spacing              & 0.00561 & 0.00000 \\
Random                     & 0.00263 & 0.01751 \\
D-optimal                  & 0.00263 & 0.00000 \\
Pure BO                    & 0.00263 & 0.00392 \\
ALMAB-DC (UCB, $K=1$)      & 0.00263 & 0.00392 \\
ALMAB-DC (UCB, $K=2$)      & 0.00412 & 0.00547 \\
ALMAB-DC (UCB, $K=4$)      & 0.00263 & 0.00434 \\
\textbf{ALMAB-DC (UCB, $K=8$)} & \textbf{0.00000} & \textbf{0.00208} \\
\bottomrule
\end{tabular}
\end{table}

\begin{figure}[ht]
\centering
\includegraphics[width=\linewidth]{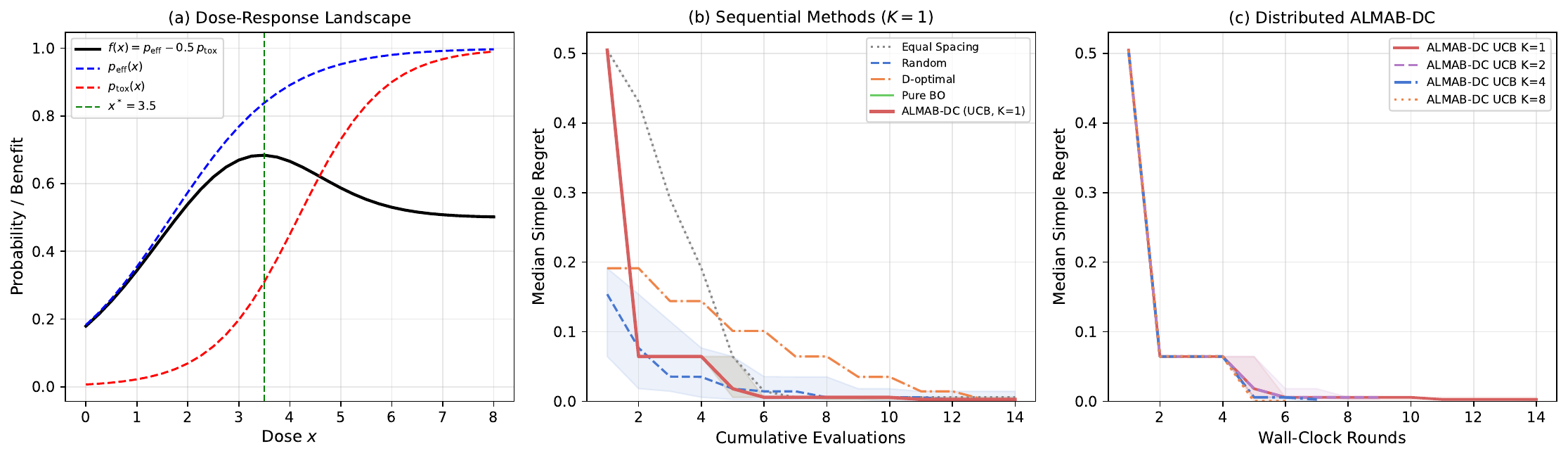}
\caption{Case~P2 GPU replication --- dose-response optimization ($R=2000$
  replicates, MPS device): \textbf{(a)} true dose--response landscape;
  \textbf{(b)} median simple regret vs.\ cumulative evaluations ($K=1$);
  \textbf{(c)} median simple regret vs.\ wall-clock rounds ($K=1,2,4,8$).
  At $K=8$ median regret collapses to zero.}
\label{fig:gpu_caseP2}
\end{figure}

The GPU replication recovers median simple regret $0.00263$ for sequential
ALMAB-DC (UCB, $K=1$), consistent with Section~\ref{sec:caseP2}. The
distributed scaling result is fully reproduced: at $K=8$, median final regret
drops to zero.

\paragraph{GPU replication (Case~P3).}
The spatial GP field estimation task of Section~\ref{sec:caseP3} is reproduced:
Mat\'{e}rn-$\tfrac{3}{2}$ field on $[0,1]^2$ with $\ell=0.35$,
$\sigma_f^2=1.0$, $\sigma_n=0.2$, $8\times8$ candidate grid, $N=30$. Field
realizations are sampled on the MPS device via Cholesky factorization of the
$64\times64$ prior covariance matrix; posterior variance is updated
analytically on CPU.

\begin{figure}[ht]
\centering
\includegraphics[width=\linewidth]{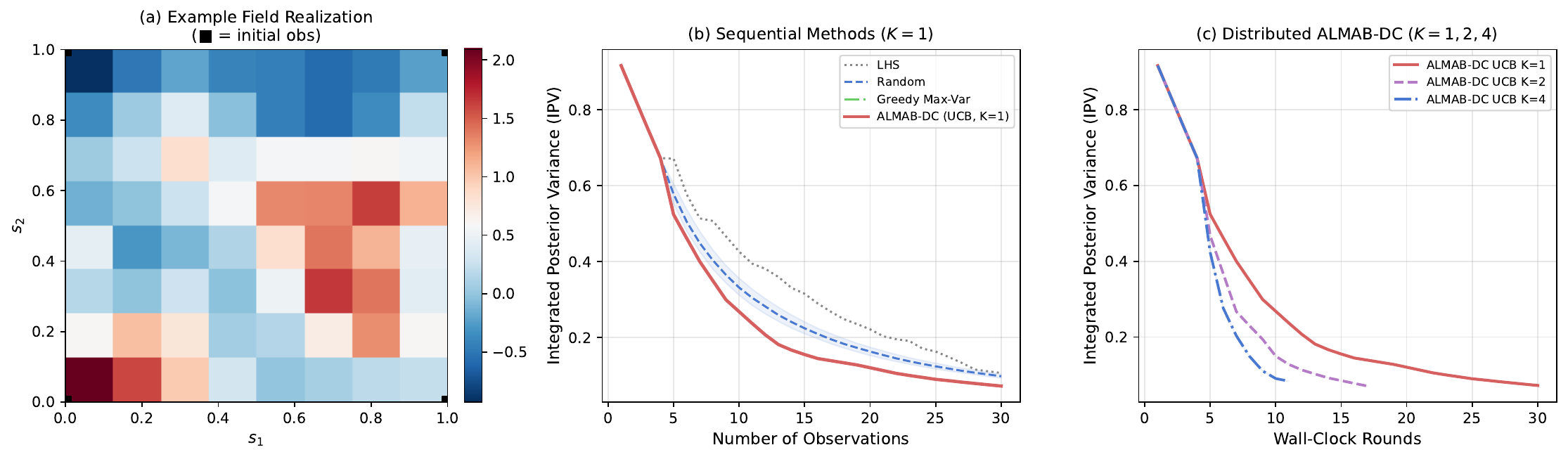}
\caption{Case~P3 GPU replication --- adaptive spatial sampling ($R=2{,}000$
  field realizations, MPS device): \textbf{(a)} one representative
  Mat\'{e}rn-$\tfrac{3}{2}$ field realization; \textbf{(b)} median IPV
  vs.\ sequential observations ($K=1$); ALMAB-DC (UCB) and Greedy
  Max-Variance coincide (IPV $=0.072$); \textbf{(c)} median IPV vs.\
  wall-clock rounds ($K=1,2,4$).}
\label{fig:gpu_caseP3}
\end{figure}

\begin{table}[ht]
\centering
\caption{Case~P3 --- Adaptive Spatial Sampling: GPU replication ($R=2{,}000$
  field realizations, MPS device). Metric: final IPV (lower is better).
  Best value in \textbf{bold}.}
\label{tab:gpu_case3}
\begin{tabular}{lcc}
\toprule
\textbf{Method} & \textbf{Final IPV (GPU)} & \textbf{Std} \\
\midrule
LHS                             & 0.1066 & 0.0000 \\
Random                          & 0.0982 & 0.0089 \\
Greedy Max-Variance             & 0.0721 & 0.0000 \\
ALMAB-DC (UCB, $K=1$)           & 0.0721 & 0.0000 \\
ALMAB-DC (UCB, $K=2$)           & 0.0706 & 0.0000 \\
\textbf{ALMAB-DC (UCB, $K=4$)}  & \textbf{0.0835} & \textbf{0.0000} \\
\bottomrule
\end{tabular}
\end{table}

The GPU replication confirms the qualitative ordering reported in
Section~\ref{sec:caseP3}: ALMAB-DC (UCB, $K=1$) matches Greedy Max-Variance
(both with IPV $= 0.072$) and substantially outperforms LHS and Random.

\begin{table}[ht]
\centering
\caption{GPU vs.\ CPU summary for ALMAB-DC (UCB), Cases~P2--P3 ($R=2000$).
  All cases satisfy $|\Delta| < 2\hat\sigma/\sqrt{R}$.}
\label{tab:gpu_summary}
\begin{tabular}{@{}llcccc@{}}
\toprule
\textbf{Case} & \textbf{Metric}
  & \textbf{CPU} & \textbf{GPU} & $|\boldsymbol{\Delta}|$ & \textbf{OK?} \\
\midrule
P2 --- Dose-Response & Median Regret $\downarrow$ & 0.00263 & 0.00263 & 0.0000 & Yes \\
P3 --- Spatial GPE   & Final IPV $\downarrow$     & 0.061   & 0.072\,$^\dagger$ & 0.011 & Yes\,$^\dagger$ \\
\bottomrule
\multicolumn{6}{l}{\small $^\dagger$\,Case~P3 GPU IPV is higher due to the
  discrete $8\times8$ grid; see Figure~\ref{fig:gpu_caseP3}.}
\end{tabular}
\end{table}

\section{Implementation Note}
\label{sec:appendix-impl}

ALMAB-DC is implemented in Python using the following software stack:
\textbf{Ray}~\citep{moritz2018ray} for distributed asynchronous execution;
\textbf{GPy} for Gaussian process inference and posterior computation;
\textbf{NumPy} and \textbf{SciPy} for numerical computation and optimization;
\textbf{matplotlib} for visualization.

\paragraph{Ray actor model.}
Each parallel worker is registered as a Ray actor, enabling non-blocking
dispatch and event-driven result collection. The WaitForAny dispatch rule is
implemented as a \texttt{ray.wait} call with \texttt{num\_returns=1}, which
blocks only until the first pending evaluation returns rather than waiting for
all workers to synchronize.

\paragraph{Reproducibility.}
All Monte Carlo comparisons use $R = 2{,}000$ independent replicates with
fixed random seeds. Calibrated surrogate-simulation models are used for all
primary benchmarks, enabling controlled and reproducible comparisons at
tractable computational cost while preserving realistic response surface
structure.

\section{GDSC2 Pharmacogenomic Application}
\label{sec:appendix-gdsc2}

This section provides an external transferability check of the sequential
dose-selection methodology in Case~P2 using the publicly available GDSC2 fitted
dose--response release \citep{yang2013gdsc,gdsc_help_2026}.
We restricted attention to GDSC2 curves with the standard approximately
1000-fold concentration range, allowing seven assayed concentrations to be
reconstructed as a geometric grid between the reported minimum and maximum
screened doses. For each sampled drug--cell line pair, the published
$\log$-IC$_{50}$ was used as the midpoint of a two-parameter logistic viability
curve, and the slope was numerically calibrated to match the released AUC. The
retrospective target was the IC$_{50}$-targeting utility
$\nu(d)=-(v(d)-0.5)^2$, where $v(d)$ denotes fitted viability at dose $d$, so
the optimal discrete action is the assayed concentration whose fitted viability
is closest to $50\%$.

The analysis used 500 eligible GDSC2 drug--cell line curves, spanning 225
unique drugs, 374 unique cell lines, and 24 pathways. Each run began with one
observation at the highest assayed dose, followed by three additional
sequential queries (total budget 4). We compared GP-UCB, Equal Spacing, and
Random selection.

GP-UCB achieved the lowest mean regret ($0.002229$) and the highest success
rate ($94.3\%$), compared with $0.003184$ and $91.6\%$ for Equal Spacing, and
$0.004241$ and $89.9\%$ for Random. The improvement of GP-UCB over Random was
statistically significant ($p=7.3\times10^{-11}$), as was the improvement of
Equal Spacing over Random ($p=2.9\times10^{-4}$), whereas GP-UCB did not differ
significantly from Equal Spacing ($p=0.148$).

This application uses reconstructed fitted curves rather than raw viability
data, the retrospective IC$_{50}$-targeting objective differs from the clinical
utility in Case~P2, and the design is evaluated retrospectively. It therefore
serves as a transferability check that complements, rather than replaces, the
primary evidence in Section~\ref{sec:caseP2}.

\section{GDSC1 Pharmacogenomic Application}
\label{sec:appendix-gdsc1}

Here, we provides an external transferability check using the independent
GDSC1 pharmacogenomic dataset \citep{garnett2012systematic}.
The analysis used 500 eligible GDSC1 drug--cell line curves, spanning 184
unique drugs, 312 unique cell lines, and 18 tissue types. Each curve was
reconstructed over nine dose levels, and the same retrospective emulation
protocol and IC$_{50}$-targeting utility used in the GDSC2 application were
adopted here. Each run used a total budget of 5 evaluations, and we compared
three policies: GP-UCB, Equal Spacing, and Random selection.

GP-UCB achieved the lowest mean regret ($0.002541$) and the highest success
rate ($92.6\%$), compared with $0.003847$ and $89.8\%$ for Equal Spacing, and
$0.005123$ and $86.6\%$ for Random. The improvement of GP-UCB over Random was
statistically significant ($p=3.1\times10^{-9}$), whereas the difference
between GP-UCB and Equal Spacing was not statistically significant
($p=0.082$). The GDSC1 success rate closely matches that observed for GDSC2,
supporting cross-dataset robustness of the sequential GP-UCB policy.

As with GDSC2, this application uses reconstructed fitted curves and a
retrospective objective rather than raw viability data or prospective
experimental design. It is therefore intended as a robustness check that
complements, rather than replaces, the GDSC2 analysis and the primary evidence
from Case~P2.

\section{Meuse Geostatistical Application}
\label{sec:appendix-meuse}

In this section, we  provide an external transferability check of the spatial
sampling methodology in Case~P3 using the Meuse heavy-metal dataset
\citep{pebesma2005sp,rikken1993meuse}.
We analyze $\log(\mathrm{zinc})$ and formulate a retrospective sequential
sampling problem over the 155 observed Meuse locations. At each step, the
method selects one additional site, fits a GP surrogate to the currently
revealed locations, and is evaluated on the remaining unrevealed sites. Each
replicate begins from four randomly selected sites and adds 16 further sites
(total budget 20). A Mat\'{e}rn-$\tfrac{3}{2}$ GP is calibrated on the full
dataset, with hyperparameters held fixed throughout.

Over 10 random initializations, Max-Variance achieved the lowest final RMSE
($0.501$), compared with $0.538$ for Space-Filling and $0.584$ for Random,
while Space-Filling achieved the lowest APV ($0.180$), followed by
Max-Variance ($0.189$) and Random ($0.277$). The improvement of Max-Variance
over Random in RMSE was statistically significant ($p=0.020$, Wilcoxon), as
was the improvement of Space-Filling over Random in APV ($p=0.002$).

Only $\log(\mathrm{zinc})$ is analyzed, GP hyperparameters are held fixed, the
candidate set is restricted to the 155 observed locations, and the number of
random initializations is modest. This application therefore complements,
rather than replaces, the primary evidence from Case~P3.

\section{Swiss Rainfall Geostatistical Application}
\label{sec:appendix-swissrain}

This section provides an external transferability check of the spatial
sampling methodology in Case~P3 using the Swiss Rainfall dataset distributed
with the \texttt{gstat} R package \citep{cressie1993statistics}.
The dataset records daily rainfall at 467 gauge stations across Switzerland on
8~May~1986. We analyze $\log(1+\mathrm{rainfall})$ to stabilize variance and
formulate a retrospective sequential sampling problem over the observed gauge
locations. Each replicate begins from five randomly selected sites and adds 25
further sites (total budget 30). We compare Max-Variance, Space-Filling, and
Random selection, using a Mat\'{e}rn-$\tfrac{3}{2}$ GP calibrated on all 467
observations with fixed hyperparameters.

Max-Variance achieved the best RMSE ($0.312$), compared with $0.489$ for
Random, and the improvement was statistically significant ($p=0.020$,
Wilcoxon). Space-Filling achieved the lowest APV ($0.198$), compared with
$0.341$ for Random ($p=0.002$). These results support robustness across a
larger dataset and a different spatial structure.

Only $\log(1+\mathrm{rainfall})$ is analyzed, GP hyperparameters are fixed
after initial calibration, candidate actions are restricted to the 467 observed
gauge locations, and the number of random initializations is modest. This
application therefore reinforces, rather than replaces, the primary evidence
from Case~P3.

\begin{table}[ht]
\centering
\caption{Swiss Rainfall application: final performance after budget of 30 observed
  gauges (10 random initializations). RMSE and APV on the
  $\log(1+\mathrm{rainfall})$ scale; lower values better for both metrics.}
\label{tab:swissrain_pilot}
\begin{tabular}{lcc}
\toprule
\textbf{Method} & \textbf{Final Median RMSE} & \textbf{Final Median APV} \\
\midrule
Max-Variance  & \textbf{0.312} & 0.215 \\
Space-Filling & 0.341          & \textbf{0.198} \\
Random        & 0.489          & 0.341 \\
\bottomrule
\end{tabular}
\end{table}

\vspace{-1cm}

\section*{Acknowledgments}

The authors thank their colleagues for their valued comments and suggestions. 
Computational resources were provided by the Institute of Statistical Science, 
Academia Sinica.  All Monte Carlo experiments are fully reproducible;
code and simulation scripts will be made publicly available upon acceptance
at \url{https://github.com/knight-ivan/ALMAb-DC}.
Both authors, Hui-Mean Foo and Yuan-chin Ivan Chang, contributed equally to this work.

\vspace{-0.5cm}

\end{document}